\pdfoutput=1

\documentclass[11pt]{article}

\usepackage[]{acl}

\usepackage{times}
\usepackage{latexsym}

\usepackage[T1]{fontenc}

\usepackage[utf8]{inputenc}

\usepackage{microtype}

\usepackage{inconsolata}

\usepackage{graphicx}

\usepackage{booktabs}
\usepackage[intlimits]{amsmath}  
\usepackage[colorinlistoftodos,size=tiny]{todonotes}
\usepackage{caption}
\usepackage{subcaption}
\usepackage{multirow}
\usepackage{enumitem}
\usepackage{makecell}

\newcommand{\bh}[1]{\textcolor{black}{#1}}
\newcommand{\sina}[1]{\textcolor{black}{#1}} 
\newcommand{\se}[1]{\textcolor{black}{#1}}
\newcommand{\hannes}[1]{\textcolor{black}{#1}} 
\newcommand{\yc}[1]{\textcolor{black}{#1}} 
\newcommand{\ycr}[1]{\textcolor{black}{#1}} 
\newcommand{\ycrr}[1]{\textcolor{black}{#1}} 
\newcommand{\yca}[1]{\textcolor{black}{#1}}
\newcommand{\senew}[1]{\textcolor{black}{#1}} 
\newcommand{\senewest}[1]{\textcolor{black}{#1}} 

\newcommand{\yccr}[1]{\textcolor{black}{#1}}

\newcommand{\quatrains}[0]{\texttt{QuaTrain}}
\newcommand{\sonnets}[0]{\texttt{SonNet}}

\newcommand{\ds}[0]{\textit{DeepSpeare}}
\newcommand{\sa}[0]{\textit{SA}}

\newcommand{\gpttwos}[0]{\textit{GPT2}$_{\textit{S}}$}
\newcommand{\gpttwosc}[0]{\textit{GPT2}$_{\textit{S}}^{\textit{con}}$}
\newcommand{\gpttwol}[0]{\textit{GPT2}$_{\textit{L}}$}
\newcommand{\gpttwolc}[0]{\textit{GPT2}$_{\textit{L}}^{\textit{con}}$}

\newcommand{\gptneos}[0]{\textit{GPTNeo}$_{\textit{S}}$}
\newcommand{\gptneosc}[0]{\textit{GPTNeo}$_{\textit{S}}^{\textit{con}}$}
\newcommand{\gptneol}[0]{\textit{GPTNeo}$_{\textit{L}}$}
\newcommand{\gptneolc}[0]{\textit{GPTNeo}$_{\textit{L}}^{\textit{con}}$}

\newcommand{\llamatwos}[0]{\textit{LLaMA2}$_{\textit{S}}$}
\newcommand{\llamatwosc}[0]{\textit{LLaMA2}$_{\textit{S}}^{\textit{con}}$}
\newcommand{\llamatwol}[0]{\textit{LLaMA2}$_{\textit{L}}$}
\newcommand{\llamatwolc}[0]{\textit{LLaMA2}$_{\textit{L}}^{\textit{con}}$}

\newcommand{\llamathr}[0]{\textit{LLaMA3}}
\newcommand{\llamathrc}[0]{\textit{LLaMA3}$^{\textit{con}}$}

\newcommand{\bygpts}[0]{\textit{ByGPT5}$_{\textit{S}}$}
\newcommand{\bygptsc}[0]{\textit{ByGPT5}$_{\textit{S}}^{\textit{con}}$}
\newcommand{\bygptl}[0]{\textit{ByGPT5}$_{\textit{L}}$}
\newcommand{\bygptlc}[0]{\textit{ByGPT5}$_{\textit{L}}^{\textit{con}}$}

\title{
Evaluating Diversity in Automatic Poetry Generation}

\author{Yanran Chen$^1$, Hannes Gröner$^2$, Sina Zarrieß$^2$, Steffen Eger$^1$ \\
 $^1$ NLLG, University of Mannheim \& University of Technology Nuremberg  (UTN); \\\url{https://nl2g.github.io/}\\$^2$ Computational Linguistics, Bielefeld University
 \\
 \texttt{yanran.chen@uni-mannheim.de}\\
  \texttt{\{hannes.groener,sina.zarriess\}@uni-bielefeld.de} \\
  \texttt{steffen.eger@utn.de}
  }

\begin{document}

\maketitle
\begin{abstract}
Natural Language Generation (NLG), and more generally generative AI, are among the currently most
impactful research fields. Creative  NLG, such as automatic poetry generation, is a fascinating niche in this area. While most previous research has focused on forms of the Turing test when evaluating automatic poetry generation --- can humans distinguish between automatic and human generated poetry --- we evaluate the \emph{diversity} of automatically generated poetry \yccr{(with a focus on quatrains)}, by comparing distributions of generated poetry to distributions of human poetry along structural, lexical, semantic and stylistic dimensions, assessing different model types (word vs.\ character-level, general purpose LLMs vs.\ poetry-specific models), \ycr{including the very recent LLaMA3\yccr{-8B}, }and types of fine-tuning (conditioned vs.\ unconditioned).
We find that current automatic poetry systems are considerably underdiverse along  
\ycr{multiple} dimensions --- they \ycr{often} 
do not rhyme sufficiently, are semantically too uniform and even do not match the length distribution of human poetry. 
\senewest{Our experiments reveal, however, that style-conditioning and character-level modeling clearly increases diversity across virtually all dimensions we explore.}
Our identified limitations may serve as the basis for more genuinely  
\senewest{diverse} 
future poetry generation models.\footnote{Code + data: \url{https://github.com/hgroener/diversity_in_poetry_generation}}  
\end{abstract}

\section{Introduction}\label{sec:intro}

\sina{A key aspect of creative language generation is the ability to create new, original and interesting text, cf.\ \cite{colton2012computational,gatt2018survey,yi2020mixpoet,Elgammal2017CANCA}.} 
\sina{
To date, 
extremely little attention has been given to the  evaluation of originality and creativity in recent creative text generation models such as those for automatic poetry generation, despite renewed interest in the context of recent LLMs \cite{franceschelli2023creativity}.}
\se{In fact, existing automatic poetry generation models are typically not evaluated regarding} \sina{how different generated poems are from existing poems in the training set }\se{
but 
with 
the \emph{Turing test}: can humans distinguish whether a poem is human authored or automatically generated \citep{hopkins-kiela-2017-automatically,lau2018deep,Manjavacas2019ARS}? 
However, this form of Turing test} \sina{and other similar forms of human evaluation} \se{may contain an} 
overlooked 
risk of 
failure: namely, if the automatically generated instances are 
(near-)copies of  
training data instances.

\vspace{-.1cm}
\par{
In this work, we fill this gap and evaluate, for the first time, \yccr{(fine-tuned)} automatic poetry generation systems in terms of their \emph{diversity}. \sina{As human evaluation is generally not well suited to assess diversity \cite{hashimoto-etal-2019-unifying}, we automatically measure diversity by comparing distributions of generated and existing poems along formal, semantic and stylistic dimensions.} \se{This yields much better evidence of the models' 
creative capabilities in contrast to being mere 
`stochastic parrots'.
}}\looseness=-1

\vspace{-.1cm}
\se{Our main contributions are: \textbf{(i)} we conceptualize diversity of poetry generation systems along different dimensions: diversity on the structural \senewest{(e.g., length)}, \senewest{stylistic (e.g., rhyming), } lexical \senewest{and} semantic 
level; \textbf{(ii)} we assess different types of automatic poetry generation systems for diversity: general purpose word\senewest{-} and character-level LLMs, both unconditioned and style-conditioned ones, on the one hand, and poetry-specific models, on the other hand; \textbf{(iii)} we evaluate each class of model for diversity across the different dimensions, by comparing the distribution of the human authored training data set to the distribution of generated poems. We find that on a distributional level, generated poems are considerably different from human ones. 
Character-level style-conditioned general-purpose LLMs 
are most diverse.} 

\se{Our work prepares the groundwork for truly creative generative AI models \citep{veale2020leaps} and also has implications for the detection of generative AI \citep{Sadasivan2023CanAT}. 
}

\vspace{-.1cm}
\section{Related Work}\label{sec:related}
\vspace{-.1cm}
Our work connects to research on diversity and automatic poetry generation, which we now discuss. 

\paragraph{Diversity} Building systems 
able to generate diverse output has been a long-standing concern in NLG research \citep{reiter-sripada-2002-squibs,van-deemter-etal-2005-squibs,foster-white-2007-avoiding} and remains a central issue in neural NLG \citep{holtzman2019curious}. 
The need for careful analysis of NLG systems' diversity -- beyond an assessment of the quality or fluency of single-best generation outputs -- has been widely acknowledged \citep{gatt2018survey,hashimoto-etal-2019-unifying,mahamood-zembrzuski-2019-hotel,celikyilmaz2020evaluation,tevet-berant-2021-evaluating,schuz-etal-2021-diversity}.
A well-known finding from this line of research is that neural NLG systems typically face a quality-diversity trade-off \citep{ippolito-etal-2019-comparison,caccia2020,wiher2022decoding}: their outputs are either well-formed and fluent or diverse and variable.

\par{
Work on evaluating diversity of NLG typically uses automatic metrics that quantify to what extent different outputs by the same system vary \citep{hashimoto-etal-2019-unifying}. 
In practice, though, evaluations of diversity in NLG differ widely across tasks \citep{tevet-berant-2021-evaluating} and even adopt different notions of diversity \citep{zarriess-etal-2021-decoding}.
At the same time, most of these notions focus on lexical or semantic aspects of diversity, e.g., \textit{local lexical diversity}.
For instance, \citet{ippolito-etal-2019-comparison} compare decoding methods in dialogue generation and image captioning, assessing lexical overlaps in $n$-best NLG outputs for the same input. 
\ycr{\citet{chakrabarty-etal-2022-help} simply measure the local lexical diversity in automatic generated poems in terms of distinct unigrams.}
\textit{Global lexical diversity}, on the other hand, measures whether the NLG system generates different outputs for different inputs. For instance, \citet{van-miltenburg-etal-2018-measuring} define the {global} diversity of image captioning systems as their ability  to generate  different captions for a set of inputs,
using metrics like the number of types in the output vocabulary, type-token ratio, and the percentage of novel descriptions. Similarly, \citet{hashimoto-etal-2019-unifying}  view diversity as related to the model's ability to generalize beyond the training set, i.e., generate novel sentences.
}\looseness=-1

Besides lexical diversity, work on open-ended or creative text generation tasks has been interested in diversity at a more general semantic level. For instance, \citet{zhang2018} and \citet{stasaski-hearst-2022-semantic} aim at building dialogue systems that generate entertaining and semantically diverse responses in chit-chat dialog. 
Here, semantic diversity has been measured, e.g., with the help of embedding-based similarity \citep{du-black-2019-boosting}.

\sina{In our work
on diversity in poetry generation, 
\emph{we complement 
\ycr{both} lexical and semantic aspects of diversity with aspects of formal diversity.}} \emph{We thus explore whether automatic poetry generation systems are able to capture the `full bandwidth' of realizations of poetry found in the data distribution with which they have been trained, focusing mostly on global diversity.}

\paragraph{Poetry generation} Automatic poetry generation is a long standing dream of AI research, dating back at least to the mid 20th century (e.g., Theo Lutz' \emph{Stochastische Texte}). While early modern systems were heavily hand-engineered \cite{gervas2001expert}, more recent approaches are all trained on collections of human poetry \cite{lau2018deep,jhamtani2019learning,agarwal-kann-2020-acrostic} but still extensively utilize human guidance e.g.\ to enforce formal characteristics of poetry such as rhyming \cite{wockener-etal-2021-end}. \citet{belouadi2022bygpt5} have recently released a character-level decoder-only LLM (ByGPT5) capable of learning style-constraints such as rhyming without human involvement in model design. 
\ycr{
\citet{chakrabarty-etal-2022-help} \ycrr{propose a collaborative system for 
\senewest{poetry,} 
which can follow human instructions to write poems. They}
measure creativity 
of the generated poems via crowd workers\senewest{,} 
\senewest{who decide} 
which of two poems is more creative. \senewest{While \citet{chakrabarty-etal-2022-help} do not define creativity, it could be considered as generating novel poems \emph{outside} the training data set; in contrast, we measure diversity 
by assessing whether poetry generation systems generate outputs that are as diverse as their human training data.}}

\emph{In our work, we explore varying poetry generation models with regard to diversity: poetry-specific models that use hand-engineered architectures as well as general purpose LLMs, including ByGPT5.}    
\section{Diversity in Poetry Generation}

We first conceptualize diversity in poetry generation using 
formal and semantic criteria.

\paragraph{Memorization.}
\par{
In poetry, as in other forms of art, 
creativity \citep{sternberg1999handbook} plays a central role. 
A basic
aspect of creativity 
is the models' ability to generate poems that are 
different from the training data, i.e.\ have not been memorized as a whole. 
To examine memorization, we proceed 
as in \citet{belouadi2022bygpt5}. 
We apply the Ratcliff-Obershelp similarity \citep{ratcliff1988pattern} 
to compare each poem in a sample with poems in the training corpus. If a generated quatrain exhibits a similarity score of 
$\ge$0.7 with a quatrain in the training data, we classify it as  
memorized. 
\yc{
A quatrain can be divided into 4 verses or 2 couplets; thus,
we also inspect memorization at the verse and couplet levels by comparing each verse or couplet in a sample to those in the training data. Higher thresholds for classification are used for these finer-grained comparison levels, as shorter texts have higher chances of being more similar in general. 
Specifically, a verse with a similarity score $\ge$0.9 or a couplet $\ge$0.8 is considered as memorized.}
We define the memorization score of a sample as the proportion of memorized quatrains in that sample. \se{How much LLMs memorize from their training data has been a question of central concern recently \cite{mccoy-etal-2023-much}.}}\looseness=-1

\paragraph{Poem length.}

\sina{Within a sample of generated poems}, we consider differences at the level of  poem length, i.e., their number of tokens, 
as a basic aspect of diversity at the formal or structural level. 
We \sina{analyze} to what extent the length distribution of generated poems differs from the distribution in the training data. 
We define the length of a quatrain as the number of tokens 
contained: 
we eliminate all punctuation symbols and split the remaining text 
by white space.
We report mean length, standard deviation, minimal and maximal length of samples. 
We additionally deploy  
distance measures 
between training data distribution and generated samples, 
in particular, 
a metric called histogram intersection \citep{swain1991color}, 
which measures the intersection area of two normalized histograms (and therefore returns values between 0 and 1). 

\paragraph{Rhyme patterns.}
As a more complex dimension of formal diversity, we consider 
rhyming as 
a  
central aspect that 
characterizes the structure of a poem. 
Diversity can then be assessed by comparing rhyme 
distributions between generated samples and training data. 
 In order to classify rhymes in our samples, we use the same 
 classifier 
used to annotate QuaTrain \citep{belouadi2022bygpt5}. 
We distinguish between true rhymes, which involve different words, 
and repetitions, which refer to rhymes based on the same word. 

\paragraph{Lexical diversity.} 
Lexical diversity is a standard aspect of diversity evaluation in NLG and is used to assess how generation outputs vary in their vocabulary, either at the local text level or at the global corpus level.
We use the following metrics to measure the lexical diversity for both the training data and the generated samples:  
(i) \textbf{Averaged type token ratio (ATTR).} We calculate ATTR 
 as the average of all type token ratios \citep{richards1987type} (TTRs) for each quatrain in a sample, i.e.\ as a measure of local lexical diversity.  
(ii)  
\textbf{Moving average type token ratio (MATTR).} The 
MATTR \citep{covington2010cutting} 
acts on the corpus level and calculates a moving average by sliding through the corpus using a window of fixed size. \bh{We deploy this metric as a measure of global lexical diversity.} 
(iii) 
\textbf{Measure of textual, lexical diversity (MTLD).} The 
MTLD \citep{mccarthy2005assessment} 
is calculated as the average length of a substring that maintains a specified TTR level. \bh{MTLD is deployed to measure lexical diversity on a global scale.}

\paragraph{Semantic diversity.} 
\par{
Even if a poetry generation system does not directly copy data from the training data, the generated poems may still be semantically very similar to the training data distribution. 
We 
employ a multilingual distilled version of Sentence-BERT (SBERT)  
\cite{reimers-gurevych-2019-sentence} 
as dense vector representations 
to measure semantic similarity between 
\se{poems: (i) across the human train set and the generated poems, (ii) within human and generated poems.} 
\se{In particular, for each generated quatrain, we note down the similarity value of the \emph{most similar} human quatrain, then report the average over all those maximum similarity values. We proceed analogously within the human training data and within the automatically generated poems}.}\looseness=-1

\section{Experiment Setup}\label{sec:setup}

\paragraph{Data}

\begin{table}
    \centering
    \resizebox{\columnwidth}{!}{%
    \begin{tabular}{@{}ccccc@{}}
\toprule
\multicolumn{1}{l}{} & \multicolumn{2}{c}{\textbf{DE}} & \multicolumn{2}{c}{\textbf{EN}} \\
\cmidrule(lr){2-3}
\cmidrule(lr){4-5} 
\multicolumn{1}{l}{} & \quatrains{}   & \sonnets{}   & \quatrains{}   & \sonnets{} \\ \cmidrule(lr){1-3}
\cmidrule(lr){4-5}
Train                 & 253,843      & 72,526     & 181,670     & 51,905     \\
Dev                   & 28,205       & 8,058      & 20,186      & 5,767      \\ \cmidrule(lr){1-3}
\cmidrule(lr){4-5}
Total                 & 282,048      & 80,584     & 201,856     & 57,672     \\ \bottomrule
\end{tabular}}
\vspace{-.2cm}
    \caption{Number of quatrains/sonnets in our datasets.}
    \label{tab:training-quatrain}
    \vspace{-.3cm}
\end{table}

We use \yc{the} QuaTrain
dataset 
published by \citet{belouadi2022bygpt5}, 
\yc{which}
consists of English and German quatrains 
\yc{from} different publicly available poetry datasets. 
\yc{The dataset }
\se{contains human written quatrains but mixes them synthetically}: 
every sequence of four consecutive lines from the underlying human 
data 
are included \se{in order to increase dataset size}.  
\yc{Besides, it} is automatically annotated for  
meter and 
rhyme 
using high-quality classifers (especially for rhyme). 
\hannes{Because our focus lies on the diversity of model outputs, we have to avoid repetitions in the training data created by the data augmentation methods used in its creation. To avoid lines appearing multiple times, we first parse the dataset sequentially, eliminating quatrains that overlap the preceding one. Because this method does not eliminate all overlaps, we then use a heuristic, deleting the ten percent of the quatrains which have the biggest overlap with other quatrains until there is no overlap remaining. \yc{We refer to the resulting dataset \se{(again)} as \quatrains{}.}} 

\yc{
\quatrains{} is split into train and dev sets using a ratio of 9:1; we do not keep a test set since no held-out human data is needed for generation or evaluation.
Further, as some models used in this work
are designed to process sonnets and/or limerick data, we create pseudo sonnets for them, denoted as \sonnets{}. Specifically, for each sonnet, we randomly draw three quatrains and one couplet from the corresponding data split of \quatrains{}, ensuring that each comes from a different original quatrain. 
Table \ref{tab:training-quatrain} provides the data sizes. 
} 

\begin{table}[!t]
\setlength{\tabcolsep}{3pt}
\resizebox{\columnwidth}{!}{
\begin{tabular}{@{}ccccc@{}}
\toprule
Class                                 & Model      & Smaller    & Larger    & Lang \\\midrule
\multirow{2}{*}{\makecell{Poetry-\\specific}}      & DeepSpeare & -          & -         & de/en     \\
                                      & SA         & -          & -         & de/en     \\ \midrule
\multirow{5}{*}{\makecell{Unconditioned\\/ Conditioned\\ LLMs}} & ByGPT5     & 140m       & 290m      & de/en     \\
                                      & GPT2       & 117m       & 774m      & de/en     \\
                                      & GPTNeo     & 125m       & 1.3b      & en        \\
                                      & LLaMA2     & 7b         & 13b       & de/en     \\
                                      & LLaMA3     & \multicolumn{2}{c}{8b} & de/en   \\\bottomrule 
\end{tabular}
}
\caption{Models used in this work. The `Smaller' and `Larger' columns display the sizes of the models considered. The `Lang' column indicates for which languages the models were trained.}
\label{tab:models}
\vspace{-.3cm}
\end{table}

\paragraph{Models}

\se{We} 
use 2 different model classes:

\begin{itemize}[topsep=2pt,itemsep=-1pt,leftmargin=*]
    \item \textbf{Poetry-specific Models}: 
    We select two models that integrate LSTM language models with additional components to generate quatrains with rhymes. \textbf{\ds{}} \citep{lau2018deep} utilizes a pentameter model to learn iambic meter and a rhyme model to distinguish between rhyming and non-rhyming words. \textbf{\textit{Structured Adversary} (\sa{})} \citep{jhamtani2019learning} learns to rhyme in an adversar\senewest{ial} setup, where a language model aims to generate poems misclassified by the discriminator, while a discriminator is trained to differentiate between generated and real poems. \emph{Both models can 
    take sonnets as input during training 
    and output quatrains during inference}. For more detailed model 
    \se{descriptions, see} 
    Appendix \ref{app:ds_sa}.
    
    \item \textbf{General Purpose LLMs}: We consider several decoder-only transformer-based models, encompassing both (sub)word- and character-level models, as well as older and very recent models.
    We choose two model families from the GPT series, GPT2 \citep{radford2019language} and GPTNeo \citep{black-etal-2022-gpt} (a replicated version of GPT3 by EleutherAI\footnote{\url{https://www.eleuther.ai/}}), two from the LLaMA series, LLaMA2 \citep{touvron2023llama} and LLaMA3 \citep{llama3modelcard}, and the \emph{character-level} ByGPT5 
    \citep{belouadi2022bygpt5}. Except for LLaMA3, we consider one smaller and one larger variant within each model family based on model size. 
    We train each model in both \textbf{unconditioned and conditioned} manners, with rhymes and meters exposed during training in the latter case. 
    \yca{We encode styles with special tokens during training and allow the models to predict the styles autonomously during inference.}
    \yc{For all LLMs, we employ consistent \textbf{decoding} strategies for generation: we use the default settings of the LLaMA2 chat models on Hugging Face\footnote{\url{https://huggingface.co/spaces/huggingface-projects/llama-2-7b-chat}} but limit the number of newly generated tokens to 100 for the word-level models} \hannes{
    \se{and} 
    300 for the character-level ByGPT5 models.}
    \end{itemize}

We end up with \yc{a total of 36} models for German and English, categorized into three groups: 1) poetry specific LSTM-based models, 2) unconditioned LLMs, and 3) conditioned LLMs, as summarized in Table \ref{tab:models}.
\sonnets{} is used for training 1), while \quatrains{} is used for 2) and 3), separately for each language.
We train all models using early stopping based on the perplexity/loss observed in the dev sets (details see Appendix \ref{app:training}), 
as overfitting may negatively bias certain metrics like memorization rates. 
\yc{To distinguish between the different sizes and training manners of the LLMs, we use the following notation:
   a subscript of S/L indicates whether it is a smaller/larger version, and a superscript of ``con'' stands for conditioned training. E.g., \gpttwos{} and \gpttwosc{} represent the unconditioned and conditioned trained GPT2 small models, respectively.}

\begin{table*}[!t]
\centering
\begin{subfigure}[b]{\textwidth}
\centering
\resizebox{.65\textwidth}{!}{
\begin{tabular}{@{}lcccccccc@{}}
\toprule
& \multicolumn{4}{c}{Memorization ($\downarrow$)} & \multicolumn{2}{c}{Length ($\uparrow$)}& \multicolumn{2}{c}{Rhyme ($\downarrow$)} \\ 
& \multicolumn{2}{c}{DE} & \multicolumn{2}{c}{EN} & DE & EN  & DE & EN\\ \cmidrule(lr){2-3}
\cmidrule(lr){4-5} 
\cmidrule(lr){6-6}
\cmidrule(lr){7-7}
\cmidrule(lr){8-8}
\cmidrule(lr){9-9}
& Couplet & Verse & Couplet & Verse  & & & &\\ \cmidrule(lr){1-5} 
\cmidrule(lr){6-7}
\cmidrule(lr){8-9}
Poetry-specific & \textbf{0.0000} & \textbf{0.006} & \textbf{0.0000} & \textbf{0.0046} & 0.752 & 0.745 & 0.992 & \textbf{0.825}\\
Character-level & \textbf{0.0000} & 0.010 & \textbf{0.0000} & 0.0087 &\textbf{0.815} & \textbf{0.813} & \textbf{0.893} & 0.895\\
Word-level & 0.0476 & 0.048 & 0.0005 & 0.0309 & 0.686 & 0.700 & 1.057 & 0.852\\
& & & & \\
Unconditioned & \textbf{0.0003} & 0.045 & 0.0006 & 0.0324 & 0.686 & 0.681& 1.107 & 0.937 \\
Conditioned & 0.0004 & \textbf{0.028} & \textbf{0.0002} & \textbf{0.0194} & \textbf{0.760} & \textbf{0.769} & \textbf{0.913} & \textbf{0.785}\\
& & & & \\
Larger & 0.0005 & \textbf{0.037} & 0.0005 & 0.0290 & 0.713 & 0.705& 1.111 & \textbf{0.861}\\
Smaller & \textbf{0.0003} & 0.039 & \textbf{0.0003} & \textbf{0.0237} & \textbf{0.726} & \textbf{0.756} & \textbf{0.931} & 0.890\\ \bottomrule
\end{tabular}
}
\caption{Structural Properties: couplet- and verse-level \textbf{memorization} rates, histogram intersection of \textbf{length} distributions between human and system-generated poems, and KL divergence between \textbf{rhyme} distributions of human and system-generated poems.}\label{tab:avg_mem}
\vspace{1cm}
\end{subfigure}\vspace{-.8cm}
\begin{subfigure}[b]{\textwidth}
    \centering
    \resizebox{.8\textwidth}{!}{
\begin{tabular}{@{}lcccccccccc@{}}
\toprule
& \multicolumn{6}{c}{Lexical ($\uparrow$)} & \multicolumn{4}{c}{Semantic ($\downarrow$)}\\
& \multicolumn{3}{c}{DE} & \multicolumn{3}{c}{EN}  &  \multicolumn{2}{c}{DE} & \multicolumn{2}{c}{EN}      \\ \cmidrule(lr){2-4}
\cmidrule(lr){5-7} 
\cmidrule(lr){8-9}
\cmidrule(lr){10-11}
& ATTR & MATTR & MTLD & ATTR & MATTR & MTLD  & Within & Across & Within & Across\\ \cmidrule(lr){1-7}
\cmidrule(lr){8-11} 
Poetry-specific    & \textbf{0.928} & \textbf{0.895} & 162.8 & \textbf{0.890} & \textbf{0.863} & \textbf{126.0} & \textbf{0.577} & \textbf{0.669} & \textbf{0.509} & \textbf{0.601}\\
Character-level & 0.915 & 0.886 & \textbf{166.7} & 0.837 & 0.818 & 83.4& 0.582 & 0.678 & 0.522 & 0.610\\
Word-level      & 0.922 & 0.874 & 114.7 & 0.871 & 0.835 & 82.7& 0.629 & 0.693 & 0.587 & 0.634\\
          &       &       &    & & &   \\
Unconditioned   & 0.919 & 0.875 & 125.9 & 0.854 & 0.818 & 75.2& \textbf{0.613} & 0.688 & 0.580 & 0.632\\
Conditioned      & \textbf{0.921} & \textbf{0.880} & \textbf{133.2} & \textbf{0.873} & \textbf{0.845} & \textbf{90.6}& 0.619 & \textbf{0.688} & \textbf{0.571} & \textbf{0.627} \\
          &       &       &     & & &  \\
Larger    & \textbf{0.932} & \textbf{0.890} & \textbf{143.9} & \textbf{0.873} & \textbf{0.837} & \textbf{84.1} & \textbf{0.613} & 0.689 & \textbf{0.571} & \textbf{0.626}\\
Smaller   & 0.902 & 0.861 & 115.6 & 0.839 & 0.814 & 74.3 & 0.623 & \textbf{0.688} & 0.577 & 0.631\\ \bottomrule
\end{tabular}
}
\caption{Lexical and Semantic Properties: \textbf{lexical} diversity metrics and `within'/`across' \textbf{similarity} scores. }\label{tab:avg_lexical_semantic}
\end{subfigure}
\caption{Average metrics for different model type aggregations. $\downarrow$ / $\uparrow$ in the brackets indicate that lower/higher values for the metrics are better, respectively. We bold the best results for each comparison.}\label{tab:avg}
\vspace{-.5cm}
\end{table*}

\section{Evaluation}
\yccr{We first report the results of diversity evaluation in \S\ref{sec:diversity_eval}, which is our main focus, followed by an examination of the relationship between diversity and overall quality through human evaluation in \S\ref{sec:quality_eval}.}

\subsection{Diversity Evaluation}\label{sec:diversity_eval}
\se{From each model
, we randomly draw 
\yc{1000} generated poems.}
\bh{Whenever we do a direct comparison between training and generated data (e.g.\ when comparing lexical diversity), we randomly draw 10 samples of size 
\yc{1000} (matching the sample size) from the train set and use mean results as representatives. We deploy this strategy to mitigate the large discrepancy in size between human data and generated poems.}

\se{We first investigate structural properties of the generated poems (repetition of instances on a surface level, length distributions, rhyming), then consider lexical and semantic properties.} 
\senewest{After discussing each dimension of diversity, we provide a brief summary that generalizes across different model classes (e.g., poetry-specific vs.\ style conditioned vs.\ unconditioned, character- vs.\ word-level, larger vs.\ smaller). These summaries are based on Table \ref{tab:avg}.}

\vspace{-.2cm}
\paragraph{Memorization}

\begin{table}[!t]
\centering
\small
\resizebox{\columnwidth}{!}{
\begin{tabular}{@{}lcccc@{}}
\toprule
                           & \multicolumn{2}{c}{\textbf{DE}}      & \multicolumn{2}{c}{\textbf{EN}}         \\ 
                  & \multicolumn{1}{c}{verse}       & couplet & \multicolumn{1}{c}{verse}       & couplet \\ \cmidrule(ll){2-3}\cmidrule(ll){4-5}
\ds         & 0.83\%                          &         & 0.83\%                          &         \\
\sa         & \textbf{0.40\%}                          &         & \textbf{0.10\%}                          &         \\ \midrule
\bygptl     & \underline{1.30\%}$^{*}$                          &         & \underline{1.23\%}$^{*}$                          &         \\
\bygpts     & \underline{1.23\%}                          &         & \underline{0.93\%}                          &         \\
\gpttwol    & \underline{6.85\%}                          & 0.10\%  & \underline{3.90\%}                          & 0.10\%  \\
\gpttwos    & \underline{8.70\%}$^{*}$                          & 0.10\%  & \underline{4.03\%}$^{*}$                          & \underline{0.10\%}  \\
\gptneol    & \multicolumn{1}{c}{-}           &         & \underline{5.60\%}$^{*}$                          & 0.05\%  \\
\gptneos    & \multicolumn{1}{c}{-}           &         & \underline{4.73\%}                          & \underline{0.10\%}$^{*}$  \\
\llamatwol  & \underline{4.65\%}                          &         & \underline{3.45\%}$^{*}$                          & \underline{0.05\%}$^{*}$  \\
\llamatwos  & \underline{5.45\%}$^{*}$                          &         & \underline{2.48\%}                          &         \\
\llamathr   & \underline{3.60\%}                          &         & \underline{2.88\%}                       & \underline{0.05\%}  \\ \midrule
\bygptlc    & 0.90\%$^{*}$                          &         & 0.58\%                          &         \\
\bygptsc    & 0.68\%                          &         & 0.75\%$^{*}$                          &         \\
\gpttwolc   & 4.38\%                          & \underline{0.15\%}$^{*}$  & 2.33\%$^{*}$                          & 0.10\%$^{*}$  \\
\gpttwosc   & 6.90\%$^{*}$                          & 0.10\%  & 2.03\%                          &         \\
\gptneolc   & \multicolumn{1}{c}{-}           &         & 3.88\%$^{*}$                          & 0.05\%$^{*}$  \\
\gptneosc   & \multicolumn{1}{c}{-}           &         & 3.50\%                          &         \\
\llamatwolc & 4.03\%$^{*}$                          & \underline{0.05\%}$^{*}$  & 2.23\%$^{*}$                          &         \\
\llamatwosc & 0.70\%                          &         & 0.55\%                          &         \\
\llamathrc  & 2.33\%                          &         & 1.65\%                          &         \\ \bottomrule
\end{tabular}}
\vspace{-.1cm}
\caption{Verse- and Couplet-level memorization rates (lower rates are better). Only non-zero entries are displayed. We underline the higher ones between the same models with different training methods, and mark those between the same models of varying sizes with $^{*}$. The best results in each dimension are bold.}
\label{tab:mem}
\vspace{-.3cm}
\end{table}

\yc{Table \ref{tab:mem} showcases the couplet- and verse level memorization rates. Since \ycr{all models exhibit zero memorization rates} 
on \textbf{quatrain-level}, 
we omit them in the table.
}

\yc{
Considering \textbf{couplet-level} memorization, 23 out of 36 models show zero memorization, while 13 models display scores between 0.05\% and 0.15\%. The poetry-specific models, \sa{} and \ds{}, as well as the character-level ByGPT5 models, exhibit no memorization; in contrast, GPT2 and GPTNeo models show the highest rates on average (up to 0.15\% for German and 0.10\% for English). 
When comparing models of the same architecture and training methods but \emph{varying sizes}, differences are found in 6 out of 14 cases. In 5 cases, larger models have 0.05\%-0.10\% higher absolute memorization scores than their smaller counterparts (the German GPT2$^{con}$ and LLaMA2$^{con}$ models, and the English GPT2$^{con}$, GPTNeo$^{con}$, LLaMA2 models); the only exception is the English GPTNeo models, where the smaller one has a 0.05\% higher memorization rate.
On the other hand, 
\emph{conditioned models mostly outperform their unconditioned counterparts}: in 4 out of 6 cases where discrepancies in memorization rates exist, the conditioned ones  
\se{exhibit} 
lower memorization rates, with absolute declines of 0.05\%-0.10\%. 
} 

\yc{In the \textbf{verse-level} evaluation, the poetry-specific models perform best overall (0.4\%-0.83\% for German and 0.1\%-0.83\% for English), followed by the ByGPT5 models (0.68\%-1.3\% for German and 0.58\%-1.23\% for English). \sa{} is the best individual model, obtaining memorization rates of 0.4\% for German and 0.1\% for English.
Again, GPT2 is worst for German, exhibiting memorization rates of 4.38\%-8.7\%, whereas, for English, GPTNeo exhibits the highest rates, ranging from 3.5\%-5.6\%. Concerning different model sizes, we again see that \emph{larger models memorize more than their smaller counterparts}: in 9 out of 14 cases, larger models show higher memorization rates, with an average absolute increase of 0.15\%. 
Here, \emph{each conditioned model 
exhibits a strictly lower memorization rate compared to its unconditioned counterpart}, with an absolute decrease of 1.47\% on average.
}

\yc{ 
\senewest{\textbf{Overall}}: 
(1) No models exhibit severe memorization issues, such as copying entire poems or large portions of poem snippets from the training data. In terms of memorization, 
(2) among model groups, the poetry-specific and character-level models are more diverse; \sa{} is the best individual one. (3) Larger models are less diverse compared to their smaller versions. (4) Conditional training enhances model diversity.}

\paragraph{Length}\label{length}

Table \ref{tab:length} \se{(appendix)} reports 
\se{statistics on the length of poems, both human and automatically generated.} 
\hannes{The mean length of human written poems is 28 in English and 24 in German. Histogram intersection values between samples generated by the models and the human written data range from 0.61 to 0.88 in German (\llamatwol{} and \sa{}) and from 0.48 to 0.92 in English (\gptneol{} and \sa{}).
\emph{While the \sa{} models fit the distribution of the human written poems the best, the character-level ByGPT5 models also perform well consistently with histogram} intersection values between 0.77 and 0.85. 
The poems generated by German \llamatwol{} and English \gptneol{} are too short and not diverse enough (in terms of standard deviation).
The poetry-specific \ds{} models do not match the human distribution very well either, with intersection values of 0.63 and 0.57 for German and English\senew{,} respectively. Here\senew{,} too, poem lengths are too 
\senew{short} 
and not diverse enough. 
\emph{Conditioned models seem to fit the training data better across the board}, the only exceptions being German \bygpts{} and English \llamatwos{}.
Figure \ref{fig:length} \yccr{(appendix)} illustrates the length distribution of human written poems, \sa{} and \gptneol{} for English.
}

\ycrr{ 
\senewest{\textbf{Overall}},  
regarding the alignment with human distributions:
(1) Character-level ByGPT5 models generally align best with human data, followed by poetry-specific models; nevertheless, the poetry-specific \sa{} is the top individual model.
(2) Style-conditional models outperform the unconditioned trained ones.
(3) Smaller models demonstrate a better fit than the larger ones.}

\paragraph{Rhyme}

\begin{figure*}[!th]
    \centering
    \begin{subfigure}[b]{0.243\textwidth}
         \centering
         \includegraphics[width=\textwidth]{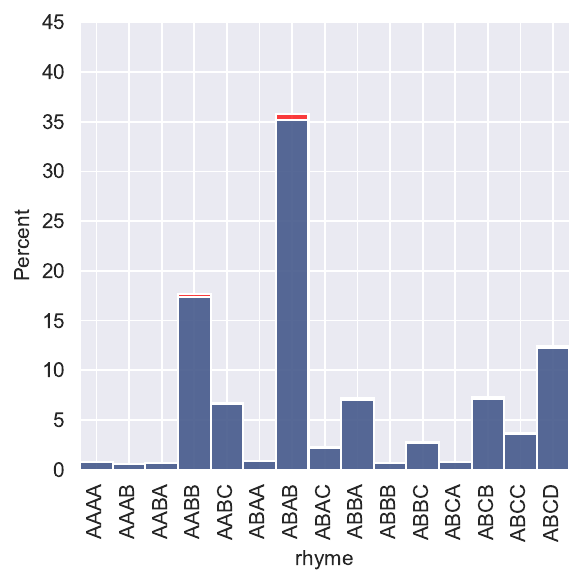}
         \caption{\textbf{Human}}
         \label{fig:r-quatrain-de}
    \end{subfigure}
    \hfill
    \begin{subfigure}[b]{0.243\textwidth}
         \centering
         \includegraphics[width=\textwidth]{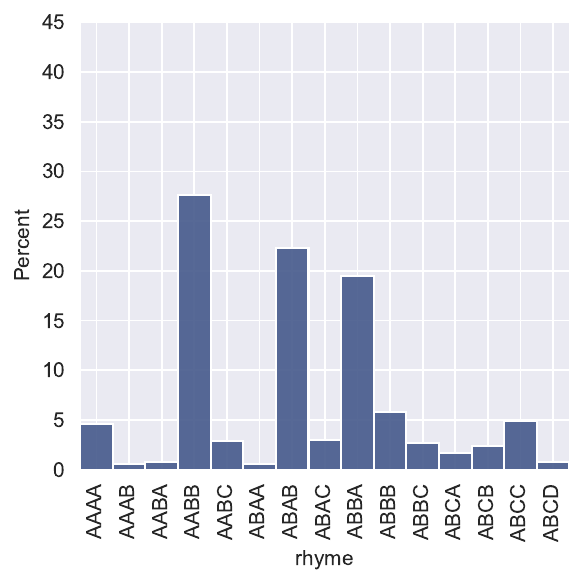}
         \caption{\textbf{Best}: \protect\ds{}}
         \label{}
    \end{subfigure}
    \hfill
     \begin{subfigure}[b]{0.243\textwidth}
         \centering
         \includegraphics[width=\textwidth]{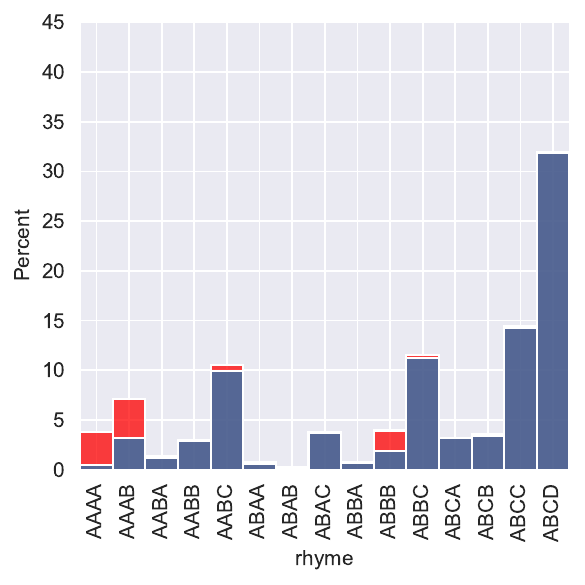}
         \caption{\textbf{Worst}: \protect\sa{}}
         \label{}
    \end{subfigure}
    \begin{subfigure}[b]{0.243\textwidth}
         \centering
         \includegraphics[width=\textwidth]{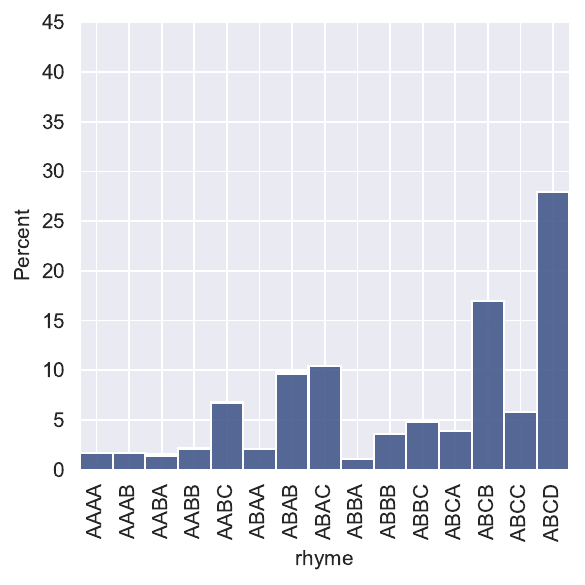}
         \caption{\textbf{Avg}: \protect\llamathr{}}
         \label{}
    \end{subfigure}
    \vspace{-.1cm}
    \caption{\yc{Distribution of rhyme schemes in (a) the human data, and the samples from the (b) best, (c) worst, and (d) average models based on their KL divergence from the human distribution for \textbf{German}.} 
    } 
    \label{fig:rhyme_de}
    \vspace{-.3cm}
\end{figure*}

\begin{figure*}[!htb]
    \centering
    \begin{subfigure}[b]{0.243\textwidth}
         \centering
         \includegraphics[width=\textwidth]{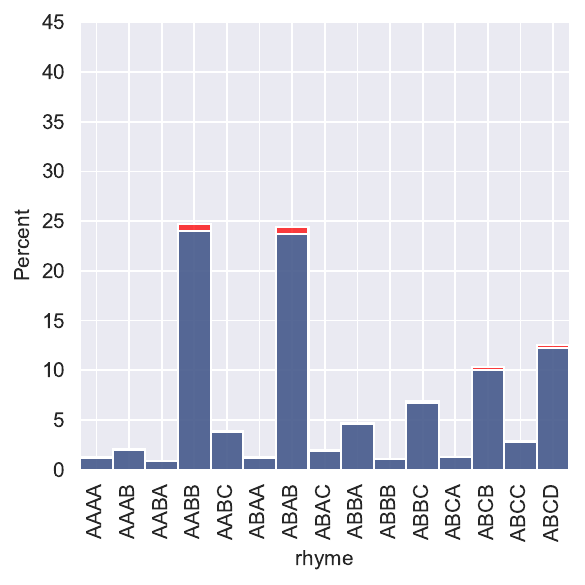}
         \caption{\textbf{Human}}
         \label{fig:r-quatrain-de}
    \end{subfigure}
    \hfill
    \begin{subfigure}[b]{0.243\textwidth}
         \centering
         \includegraphics[width=\textwidth]{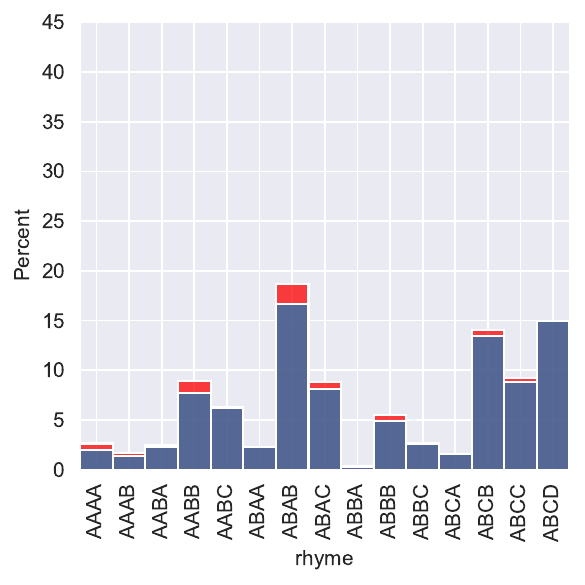}
         \caption{\textbf{Best}: \protect\gptneolc{}}
         \label{}
    \end{subfigure}
    \hfill
     \begin{subfigure}[b]{0.243\textwidth}
         \centering
         \includegraphics[width=\textwidth]{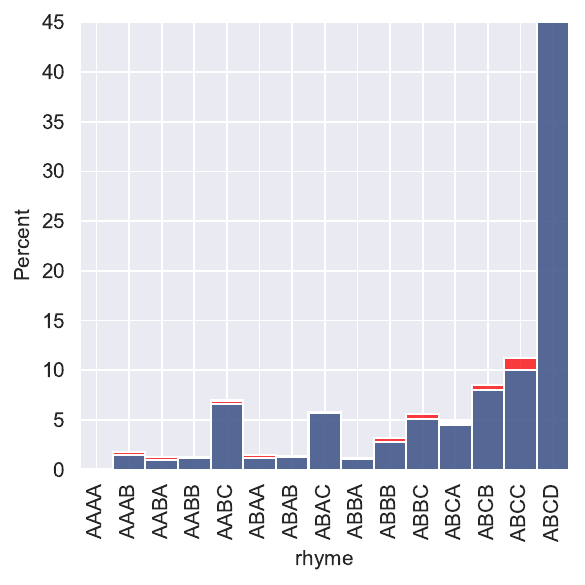}
         \caption{\textbf{Worst}: \protect\gptneol{}}
         \label{}
    \end{subfigure}
    \begin{subfigure}[b]{0.243\textwidth}
         \centering
         \includegraphics[width=\textwidth]{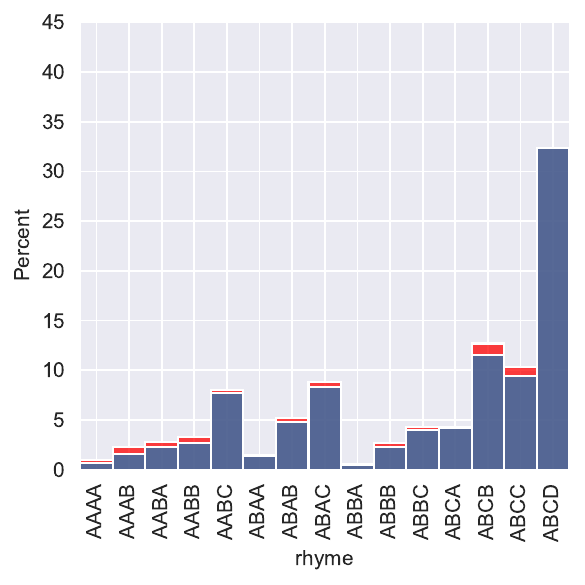}
         \caption{\textbf{Avg}: \protect\gpttwosc{}}
         \label{}
    \end{subfigure}
    \vspace{-.1cm}
    \caption{\yc{Distribution of rhyme schemes in (a) the human data, and the samples from the (b) best, (c) worst, and (d) average models based on their KL divergence from the human distribution for \textbf{English}.} 
    } 
    \label{fig:rhyme_en}
    \vspace{-.3cm}
\end{figure*}

Figures \ref{fig:rhyme_de} (a) and \ref{fig:rhyme_en} (a)  
show the distributions of rhyme schemes in our human training datasets \senew{for German and English, respectively}. 
For both languages, 
\senew{less than 15\%}
of all quatrains 
in training do not rhyme at all (rhyme scheme ABCD). Excluding ABCD, the top 3 dominant rhyme schemes by appearance are ABAB, AABB and 
\senew{ABCB}
for both datasets, with a total share of approximately 
\se{60\%}
in each language. 
\senew{German has a higher proportion of ABAB (above 35\%), while English has ABAB and AABB in roughly equal proportions (25\%).}
\senew{
Table \ref{tab:entropy} \yc{(appendix)} 
report\senew{s} the entropy of all rhyme distributions 
\senew{and} 
the distance between the human distribution and model distributions, measured in 
KL divergence. The best, worst and an average model, in terms of 
KL divergence, are shown in Figures \ref{fig:rhyme_de} and \ref{fig:rhyme_en}.}

\textbf{Poetry-specific models}: 
Figure \ref{fig:r-ds-sa} 
\se{(appendix)} shows the distributional plots for \ds{} 
and \sa{}. 
\senew{We observe that \ds{} has a very low ratio of ABCD, considerably lower than human poems (less than 5\% for both languages). The three dominating patterns are AABB, ABAB, and ABBA which (only) partially agrees with the dominating patterns in the human data. Nonetheless, \ds{} has the best fit of all models in terms of KL divergence, \yc{ranking first for German and second for English.} 
\sa{} has a much worse fit and produces considerably too many ABCD patterns (close to or above 30\% in both languages). It has one of the worst fits to the human rhyme distributions across all models.}

Figures \ref{fig:r-nonpoetry-de} and \ref{fig:r-nonpoetry-en} \se{(appendix)} 
show the distributions of rhyme patterns for \textbf{unconditioned LLMs}. 
\senew{Except for \llamathr{}, all models of this kind have a high distribution of ABCD and consequently a high likelihood of producing non-rhyming poems. Thus, they have the worst fit to the human distribution, on average, among all model classes considered.}

\textbf{Style-conditioned LLMs} are shown in 
Figures \ref{fig:r-poetry-de} and \ref{fig:r-poetry-en} \se{(appendix)}. 
\senew{In general, this model class matches the human distribution closest in terms of KL divergence. 
\senew{However,} no model produces a lot of AABB rhyme pattern which abound in our human training data. 
Across all models in this class, the fit to the human data is still mediocre at best. 
}

\par{
\senew{\textbf{Overall}, 
most models have clearly higher ABCD rhyming schemes than the human data, thus are  underdiverse concerning rhyming. 
(1) Conditioned models very clearly outperform unconditioned models and (2) character-level and poetry-specifc models are clearly better than word-level models in terms of matching the human rhyme distribution. (3) There is no clear size effect.
}}
\looseness=-1

\begin{table}[t]
    \centering
    \small
    \resizebox{\columnwidth}{!}{%
    \begin{tabular}{@{}lrrr@{}}
\toprule
Model      & ATTR (\%)   & MATTR (\%)  & \multicolumn{1}{c}{MTLD}                   \\ \midrule
HUMAN      & 91.6 / 87.7  & 90.6 / 87.3 & \underline{283.1} / \underline{183.4} \\ \midrule
\ds{}        & 92.6 / 89.1 & 87.9 / 84.8 & 110.0 / 89.7           \\
\sa{}        & 93.0 / 88.9 & \textbf{\underline{91.0}} / 87.8 & \textbf{215.6} / \textbf{162.2}          \\ \midrule
\bygpts{} & 89.7 / 81.5 & 86.9 / 79.7 & 135.4 / 66.5           \\
\bygptl{} & 91.2 / 82.5 & 88.1 / 80.5 & 151.6 / 69.9           \\
\gpttwos{}     & 86.2 / 79.4 & 81.2 / 76.4 & 64.1 / 46.0            \\
\gpttwol{}    & 94.2 / 87.6 & 89.5 / 83.5 & 131.8 / 81.6           \\
\gptneos{}     & - / 78.3    & - / 74.9    & - / 40.1               \\
\gptneol{}     & - / 86.8    & - / 81.3    & - / 61.7               \\
\llamatwos{}     & 92.8 / 89.6 & 87.7 / 86.8 & 120.7 / 106.8          \\
\llamatwol{}     & \textbf{\underline{94.8}} / 90.2 & 90.2 / 85.7 & 150.1 / 96.0           \\
\llamathr{}     & 94.4 / \textbf{\underline{92.7}} & 89.3 / 87.4 & 128.0 / 108.1          \\ \midrule
\bygptsc{} & 92.2 / 85.1 & 89.5 / 83.1 & 187.1 / 94.6           \\
\bygptlc{} & 93.0 / 85.9 & 90.0 / 83.9 & 192.6 / 102.5          \\
\gpttwosc{}     & 89.2 / 84.0 & 84.2 / 81.9 & 82.0 / 70.3            \\
\gpttwolc{}     & 94.2 / 88.0 & 90.0 / 85.3 & 137.4 / 90.7           \\
\gptneosc{}     & - / 83.1    & - / 80.2    & - / 61.2               \\
\gptneolc{}     & - / 87.0    & - / 82.1    & - / 69.4               \\
\llamatwosc{}    & 91.1 / 90.0 & 86.8 / 88.2 & 104.4 / 109.3          \\
\llamatwolc{}     & 91.9 / 90.8 & 86.5 / 87.2 & 100.2 / 101.0          \\
\llamathrc{}     & 93.5 / 91.7 & 89.1 / \textbf{\underline{88.3}} & 128.5 / 116.3          \\ \bottomrule
\end{tabular}
}
\vspace{-.1cm}
    \caption{Lexical diversity metrics for German (first entry) and English (second entry) models. \yc{Best results in each dimension are underlined; best among models are in bold.}}
    \label{tab:lexical}
    \vspace{-.3cm}
\end{table}

\paragraph{Lexical Diversity.} 

\sina{Table \ref{tab:lexical} shows the lexical diversity results for English and German.
\emph{For \textbf{local diversity (ATTR)}, most of the models are close to the diversity in human-written poems}, with the traditional models (\ds{}, \sa{}) 
and the LLaMA exceeding the ATTR values of human-written poems. For \ycr{German}, the least locally diverse poems are generated by \gpttwos{}, in the un/conditioned case, respectively. For 
\ycr{English}, the least locally diverse models is \gptneos{}, in the un/conditioned case, respectively.}  \sina{The \textbf{global diversity} metrics (MATTR, MTLD) show different trends than ATTR, though. The \textbf{MATTR} metric suggests that \emph{most models do not generally achieve the level of diversity found in human poems}: in English, only SA matches and slightly exceeds human diversity, in German, only 
the \llamatwosc{} and \llamathrc{} model exceeds human diversity.
According to the \textbf{MTLD} metric, \emph{all models generate severely under-diverse output at the sample level}. Here, the best model in English and German is \sa{}, but even \sa{} does not come close to the human level of global diversity. 
According to MTLD, \emph{style-conditioned LLMs consistently outperform their non-conditioned counterparts}, with the English LlaMA2 models being the only exceptions here. 
Moreover, we observe that model size affects all three lexical diversity metrics, whereby \emph{larger models are more diverse than their smaller counterparts.} 
The effect of size is most pronounced for GPT2, 
where ATTR, MATTR and MTLD substantially improve from the small to the larger model variant. 
Generally, the MTLD results suggest more pronounced differences between models as well as humans and models than MATTR. 
} 

\ycrr{
\textbf{
\senewest{Overall,}}   
in terms of lexical diversity, (1) neural models match human performance at the local level but fall short at the global level.
(2) Poetry-specific models outperform other model classes, while character-level LLMs are most deficient \senewest{(except for 
MTLD)}.
(3) Conditional training is beneficial.
(4) Larger models perform better.
}

\paragraph{Semantic Similarity}

\begin{table}
    \centering
    \small
    \begin{tabular}{@{}lrr@{}}
\toprule
Model      & Within (\%) & Across (\%) \\ \midrule
HUMAN      & \textbf{55.0} / \textbf{48.2}   & -           \\ \midrule
\ds{}  & 59.5 / 52.2 & 67.8 / 60.8 \\
\sa{}        & 55.8 / 49.6 & \textbf{65.9} / \textbf{59.4} \\ \midrule
\bygpts{} & 58.4 / 53.2 & 68.1 / 61.5 \\
\bygptl{} & 58.2 / 52.7 & 67.9 / 61.6 \\
\gpttwos{}     & 64.5 / 59.5 & 69.3 / 63.9 \\
\gpttwol{}      & 63.6 / 57.6 & 70.1 / 63.3 \\
\gptneos{}      & - / 62.2    & - / 63.8    \\
\gptneol{}     & - / 60.9    & - / 63.9    \\
\llamatwos{}     & 61.0 / 59.4   & 68.5 / 64.2 \\
\llamatwol{}     & 62.3 / 58.0   & 68.9 / 62.9 \\
\llamathr{}      & 61.2 / 58.4 & 69.1 / 63.8 \\ \midrule
\bygptsc{} & 58.4 / 52.2 & 67.7 / 60.8 \\
\bygptlc{} & 57.9 / 50.9 & 67.6 / 60.3 \\
\gpttwosc{}     & 64.3 / 59.2 & 70.1 / 64.3 \\
\gpttwolc{}      & 62.6 / 57.4 & 69.7 / 63.1 \\
\gptneosc{}      & - / 58.9    & - / 64.0      \\
\gptneolc{}     & - / 60.3    & - / 62.9    \\
\llamatwosc{}     & 66.9 / 57.3 & 69.3 / 64.0   \\
\llamatwolc{}      & 63.3 / 58.5 & 69.5 / 62.9 \\
\llamathrc{}    & 59.6 / 58.2 & 68.0 / 62.3   \\ \bottomrule
\end{tabular}
\vspace{-.1cm}
    \caption{ \label{tab:sem2} 
    Average 
    maximum semantic similarity values \yc{for German (first entry) and English (second entry)}: (i) within models including the training data (left) and (ii) 
    across models and humans (middle). \yc{We bold the best result in each dimension \sina{(Lower similarity means higher/better diversity)}.} 
    }
    \vspace{-.3cm}
\end{table}

\sina{Table \ref{tab:sem2} presents results for the semantic (cosine) similarity of quatrains: (i) within human and model-generated samples, and (ii) across generated samples and the human data. 
\emph{None of the models generates a sample of poems with a within-sample diversity as low as the human with-sample diversity}. \sa{} is the model that achieves the lowest within-sample similarity and the lowest across-sample similarity. 
}

\senewest{\textbf{Overall}, (1) poetry-specific models are most diverse regarding semantic similarity and word-level models are least diverse; (2) style-conditioning makes models slightly more diverse semantically; (3) larger models are also slightly more diverse.}

\paragraph{Which is the most diverse model?}
We have seen that unconditioned LLMs exhibit poor results 
\ycr{across various} dimensions of diversity: they \ycr{often} do not rhyme, are lexically underdiverse and do not show sufficient semantic variation. However, character-level models are more diverse than word level models. Style-conditioned models perform better regarding \ycr{memorization,} rhyming, 
and lexical variation, while deviating less from human poems according to the distribution match of length and rhymes.
\ycr{On the other hand, larger LLMs often outperform their smaller counterparts in semantic and lexical diversity, but they also tend to memorize more from the training data.}
\emph{Character-level style-conditioned LLMs produce overall best diversity results} and do not deteriorate as a function of 
model/training data size. \ycr{In Appendix \ref{app:best}, we calculate the average ranks of the models across all 5 dimensions, finding that indeed, for both languages, the conditioned trained ByGPT5 models perform overall best among all models, ranking as the first and second places for German and the first and third places for English.
In terms of diversity, poetry-specific 
\sa{} and \ds{} 
\ycr{overall lag only slightly behind character-level LLMs} but require 
more modeling effort from human experts (e.g., in developing rhyming components). The largest word-level LLMs explored in this work, LLaMA2 and LLaMA3, generally perform best among the word-level models; however, they do not exhibit superiority over the style-conditioned character-level models and poetry-specific models as well.
}

\ycrr{
We also compute Pearson's correlations between ranks for different dimensions. For German, the highest correlation is between semantic diversity and memorization (0.842), followed by the two moderate to high correlations: 0.526 (semantic vs.\ lexical) and 0.518 (memorization vs.\ rhyme). Two pairs show moderate correlations: 0.480 (semantics vs.\ length) and 0.404 (memorization vs.\ rhyme). The remaining pairs exhibit weak positive or negative correlations, with absolute values between 0.051 and 0.228. For English, no pairs exhibit high correlations. Two pairs show moderate to high correlations: 0.628 and 0.635 (memorization vs.\ semantics/length). Three pairs demonstrate moderate correlations, ranging from 0.307 to 0.357 (semantics vs.\ lexical/length and memorization vs.\  length). The others show weak correlations, with absolute values between 0.024 and 0.267. 
\senewest{Concretely, these sometimes low correlations are mirrored in the different ranks models have across different dimensions: for example, \sa{} is almost as diverse as the human training data regarding semantics and length, but provides one of the worst fits regarding rhyming. This indicates that most current models face a tradeoff for different diversity dimensions.}
}

\subsection{Quality Evaluation}\label{sec:quality_eval}

\yccr{Diversity in model outputs could sometimes result from low coherence or a lack of meaningful content. To investigate whether this is the case, we conducted a small-scale human evaluation of the overall quality of quatrains, focusing specifically on coherence and semantics (punctuation was omitted here, as it was also excluded during the diversity evaluation). In this evaluation, we compared 60 outputs across 5 systems (12 outputs per system) for each language, including human-written quatrains, and the outputs of the winning models in overall, lexical, semantic, and rhyme diversity (as presented in Tables \ref{tab:rank_de} and \ref{tab:rank_en} in the appendix). 
We created 15 annotation instances; in each instance, an annotator was given 4 quatrains and asked to select both the best and the worst among them.}

\yccr{
As Table \ref{tab:quality_eval} in the appendix displays, for German, human quatrains are clearly preferred (they were chosen as the best 12 times and the worst 0 times). The best automatic system is the overall winning ByGPT5 model (best 2 times; worst 1 time); \sa{} is the worst (worst 8 times). For English, the lexical winning LLaMA3 model is the best in terms of coherence (best 6 times; worst 0 times), followed by the rhyme winning GPTNEO model (best 5 times; worst 0 times); \sa{} is again the worst (worst 11 times). However, we noted that our evaluator was a native speaker of German but not English and said that the German evaluation was much easier for him. 
The older \sa{} model appears to have higher diversity at the cost of quatrain quality. However, \textbf{overall}, we conclude that more diverse models also seem to be qualitatively better --- this does not have to be a causal/strong relationship, however, especially for the newer LLMs. Tables \ref{tab:quatrain_example_de} and \ref{tab:quatrain_example_en} in the appendix present 10 sample quatrains selected as the best in our human evaluation, including both system-generated and human-written ones.
}

\section{Conclusion}\label{sec:conclusion}

\ycr{Our work is the first and most comprehensive automatic evaluation of poetry diversity, yielding several interesting observations: \senewest{for example, we find that style-conditioning enhances virtually all measures of diversity and that character-level modeling also increases diversity,  including reducing memorization.}} 
\sina{
Our evaluations \senewest{also} shed light on the fact that none of the state-of-the-art poetry generators is able to match the level of diversity in human poems. 
\senewest{Thus, we find overall} 
that an automatic assessment of the diversity of generated poems covers an important blind spot of existing studies. \senewest{Future work should aim for more diverse automatic poetry generation systems as a prerequisite of general computational creativity.}
}


\section*{Limitations}

\sina{Our work evaluates a range of existing state-of-the-art approaches, such as poetry-specific models like Deepspeare or pretrained LLMs. These models differ in various ways, with respect to their architecture, training scheme, pretraining, and the type of data they expect during training and/or finetuning. In light of these differences, it is difficult to isolate exactly how different aspects of a poetry generator impact on the  diversity of its outputs. 
\se{While our work investigated the influence of the model architecture on a high level (character vs.\ word), further aspects --- and in particular pre-training --- may be worth investigating in future work.}
}

\yc{
\ycr{Due to the hardware constraints and time limitations, we did not run experiments multiple times to take the averages or optimize the training hyperparameters, which may have introduced a degree of randomness in our results.
}
\ycrr{For example,} in our initial experiments, we trained GPT2 models with a slightly different setting.
Compared to the GPT2 models we mainly reported, these models behave slightly differently. E.g., they exhibit better lexical diversity, as shown by an increase in ATTR from 0.87 to 0.89, MATTR from 0.84 to 0.86, and MTLD from 88 to 101 on average. Similarly, they are also more diverse according to the semantic similarity metrics, which are on average $\sim$0.02-0.03 lower. In contrast, these models perform worse in rhyming; they have a $\sim$10\% lower chance of producing rhymed quatrains, and their rhyme distributions are more distant from human distributions (0.27 higher KL divergence). 
Despite these differences, our findings are generally robust \senewest{as we report averages over model classes in our analysis}. 
}
\yccr{For the same reason, we did not select the largest versions of these models; nevertheless, our evaluation already shows prominent differences in diversity across model sizes.}

\yca{
Further, we note that our trained LLMs occasionally do not generate texts in the form of a quatrain (i.e., 4 verses). These outputs were excluded from the analysis, though such cases are rare (1.5\% on average).
}

\section*{Ethics Statement}

\sina{
All the datasets, models, and code used in this work will be made publicly available. We have not collected private or sensitive data and have only used language models with free access, such that our experiments can be fully  replicated by anyone.}

\sina{Generally, our work is concerned with the evaluation of NLG systems; evaluation methods and evaluation metrics \citep{zhao-etal-2019-moverscore,Zhang2020BERTScore,peyrard-etal-2021-better,NEURIPS2021_e4d2b6e6,chen-etal-2022-reproducibility,chen_menli:2023,leiter-etal-2023-eval4nlp} are a well-known and notorious issue in this research field. 
While a lot of recent work has aimed at improving common practices in human evaluation \citep{belz-etal-2023-missing} or advancing the study of metrics for quality or fluency of NLG outputs, the evaluation of diversity is comparatively under-researched. In this work, we aimed at providing a range of metrics assessing different aspects of diversity, but could not cover all potentially interesting ways of measuring diversity. Here, future work could look at further aspects of formal and structural diversity (e.g.\ at the level of syntax, or meter), or other aspects of semantic diversity (e.g. topical diversity, rhetorical figures).} \se{Future work could also consider more (diverse) languages and other genres and datasets for poetry.}

\bibliography{custom}


\appendix
\section{Appendix}

\subsection{\ds{} and \sa{}}\label{app:ds_sa}

\textbf{Deepspeare} 
\cite{lau2018deep} 
is specifically designed for poetry generation. Its core architecture consists of an LSTM language model, a pentameter model (specifically designed to learn iambic meter) and a rhyme model. During training, it takes sonnets as input data (three quatrains followed by a couplet) but ultimately processes the contained quatrains by splitting any given sonnet.  
The rhyme model processes ending words of quatrain verses and uses a margin-based loss to discriminate between rhyming and non-rhyming words. It is not limited to specific rhyme patterns but assumes that rhymes exist in the data. 
At inference time, Deepspeare generates quatrains.

\paragraph{Structured Adversary.}
 
Like Deepspeare, Structured Adversary (SA) \cite{jhamtani2019learning} 
incorporates different components: an LSTM language model 
and a discriminator 
used to decide whether line endings 
are typical for poetry. 
Both components are organized in an adversarial setup, where the language model acts as a generator, trying to generate poems that are misclassified by the discriminator, while the discriminator is trained to distinguish generated poems from real ones. 
SA is trained with sonnets as input data. 
At inference time, it generates quatrains.

\subsection{Training}\label{app:training}

\paragraph{DeepSpeare}
\ds{} \citep{lau2018deep} 
leverages 
pretrained static word vectors. 
We
\yc{use \quatrains{} and \sonnets{} to train our own Word2vec embeddings \citep{mikolov2013efficient} and the final sonnet models respectively.} 
\yc{For the sonnet model training, we use a batch size of 128 and apply early stopping with a patience of 5 epochs; default settings are maintained for the other hyperparameters.}

\paragraph{SA}
We 
use the same word vectors and training data splits 
as for \ds{}. 
Training \sa{} 
involves 
1) pretraining the discriminator's encoder 
using a publicly available pronouncing dictionary; 
2) training the LM component; 
3) training a final aggregated model in a generative adversarial setup. 
\yc{We train the discriminators with a batch size of 128, the LMs with a batch size of 64, and the final sonnet models with a batch size of 128; here, we also implement early stopping with a patience of 5 epochs.}

\paragraph{Style-un/conditioned LLMs}

We train all LLMs for 50 epochs on our train set using the
paged AdamW optimizer with a weight decay of 0.001, a 
learning rate of 4e-05, a cosine
learning rate decay with a 3\% warmup ratio, and early stopping with patience of 5 epochs. As we run experiments on GPUs with varying memory capacities ranging from 12GB to 80GB, and with models that drastically differ in size, to achieve as much consistency as possible, we either train models with a batch size of 128 or accumulate the batches to reach a size of 128.
For LLaMA, we use 4-bit quantization and LORA \citep{hu2021lora}; the corresponding parameters are list below:
\begin{itemize}[topsep=2pt,itemsep=-1pt,leftmargin=*]
    \item target modules: q\_proj, v\_proj, k\_proj, o\_proj, embedded\_tokens
    \item lora alpha: 16
    \item lora dropout: 0.05
    \item r: 16
\end{itemize}

\subsection{Evaluation Results}

\begin{table*}[!h]
\centering
\begin{tabular}{llrrrrrr}
\toprule
$L$ & model & $h$ & $m$ & $M$ & $\mu$ & $\sigma$ & $std$ \\
\midrule
de & HUMAN & 1.00 & 4 & 65 & 24.40 & 23 & 6.39 \\
de & \ds{} & 0.63 & 14 & 30 & 21.69 & 22 & 2.45 \\
de & \sa{}  & 0.88 & 10 & 44 & 24.44 & 24 & 5.36 \\
de & \bygpts{} & 0.84 & 9 & 43 & 22.11 & 22 & 4.86 \\
de & \bygptl{} & 0.79 & 9 & 40 & 21.09 & 21 & 4.59 \\
de & \gpttwos{} & 0.59 & 9 & 32 & 19.18 & 19 & 3.54 \\
de & \gpttwol{} & 0.73 & 13 & 41 & 21.98 & 22 & 3.55 \\
de & \llamatwos{} & 0.57 & 9 & 31 & 18.84 & 19 & 3.29 \\
de & \llamatwol{} & 0.55 & 9 & 30 & 18.73 & 19 & 3.17 \\
de & \llamathr{} & 0.74 & 12 & 40 & 21.39 & 21 & 3.99 \\
de & \bygptsc{} & 0.82 & 11 & 47 & 22.38 & 22 & 4.98 \\
de & \bygptlc{} & 0.81 & 9 & 45 & 21.78 & 21 & 5.17 \\
de & \gpttwosc{} & 0.70 & 11 & 37 & 20.68 & 20 & 3.56 \\
de & \gpttwolc{} & 0.79 & 14 & 45 & 24.14 & 24 & 4.38 \\
de & \llamatwosc{} & 0.83 & 12 & 49 & 24.22 & 23 & 5.41 \\
de & \llamatwolc{} & 0.62 & 12 & 34 & 20.18 & 20 & 2.84 \\
de & \llamathrc{} & 0.76 & 10 & 47 & 21.69 & 21 & 4.14 \\
\midrule
en & HUMAN & 1.00 & 4 & 67 & 28.06 & 28 & 6.26 \\
en & \ds{} & 0.57 & 15 & 33 & 23.85 & 24 & 2.85 \\
en & \sa{}  & 0.92 & 12 & 52 & 27.36 & 27 & 5.38 \\
en & \bygpts{} & 0.80 & 12 & 44 & 25.30 & 25 & 5.09 \\
en & \bygptl{} & 0.77 & 11 & 47 & 24.97 & 25 & 4.87 \\
en & \gpttwos{} & 0.69 & 13 & 55 & 24.11 & 24 & 4.48 \\
en & \gpttwol{} & 0.72 & 13 & 56 & 24.74 & 24 & 4.94 \\
en & \gptneos{} & 0.55 & 11 & 55 & 22.67 & 22 & 3.89 \\
en & \gptneol{} & 0.48 & 13 & 34 & 21.93 & 22 & 3.16 \\
en & \llamatwos{} & 0.87 & 15 & 75 & 28.60 & 27 & 7.52 \\
en & \llamatwol{} & 0.67 & 12 & 54 & 23.95 & 24 & 4.50 \\
en & \llamathr{} & 0.59 & 14 & 60 & 23.20 & 23 & 4.23 \\
en & \bygptsc{} & 0.85 & 13 & 42 & 26.21 & 26 & 4.96 \\
en & \bygptlc{} & 0.84 & 14 & 42 & 25.85 & 25 & 4.84 \\
en & \gpttwosc{} & 0.86 & 17 & 61 & 28.37 & 27 & 6.18 \\
en & \gpttwolc{} & 0.83 & 16 & 70 & 27.82 & 27 & 6.15 \\
en & \gptneosc{} & 0.74 & 16 & 49 & 25.13 & 24 & 4.47 \\
en & \gptneolc{} & 0.53 & 12 & 35 & 22.26 & 22 & 3.36 \\
en & \llamatwosc{} & 0.70 & 17 & 74 & 33.55 & 32 & 7.83 \\
en & \llamatwolc{} & 0.81 & 15 & 56 & 26.92 & 26 & 5.80 \\
en & \llamathrc{} & 0.78 & 16 & 65 & 27.12 & 26 & 5.35 \\
\bottomrule
\end{tabular}
\caption{Reported statistical and distance measures regarding the length of training data and generated quatrains. $h$ = histogram intersection score between sample and training data, $\mu$ = mean length, $\sigma$ = median, $std$ = standard deviation, $m$ = minimal length, $M$ = maximal length.}
\label{tab:length}
\end{table*}

\begin{figure*}[t]
    \centering
    \begin{subfigure}[b]{0.32\textwidth}
         \centering
         \includegraphics[width=\textwidth]{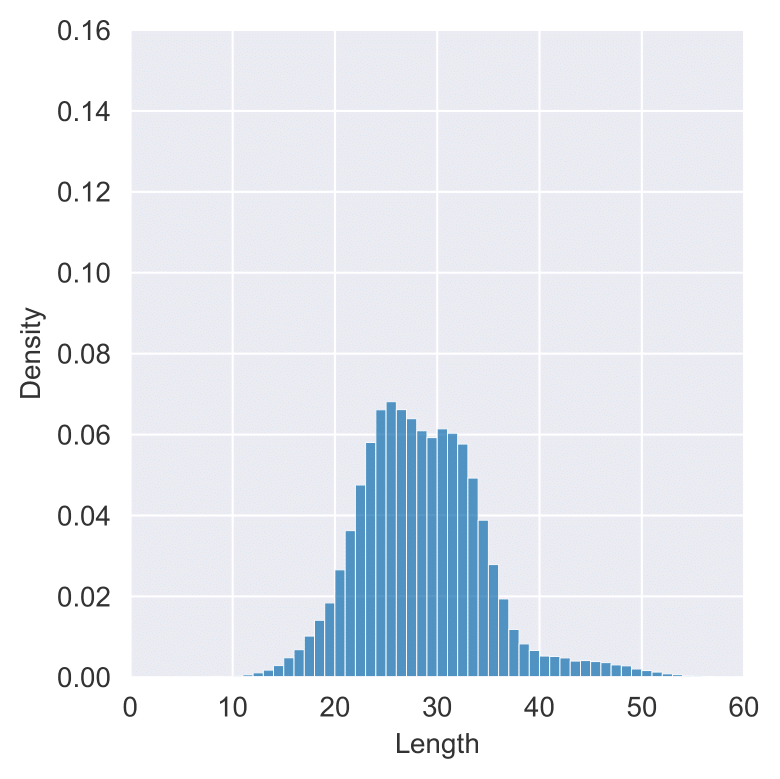}
         \caption{Human 
         }
         \label{fig:l-quatrain-en}
    \end{subfigure}
    \begin{subfigure}[b]{0.32\textwidth}
         \centering
         \includegraphics[width=\textwidth]{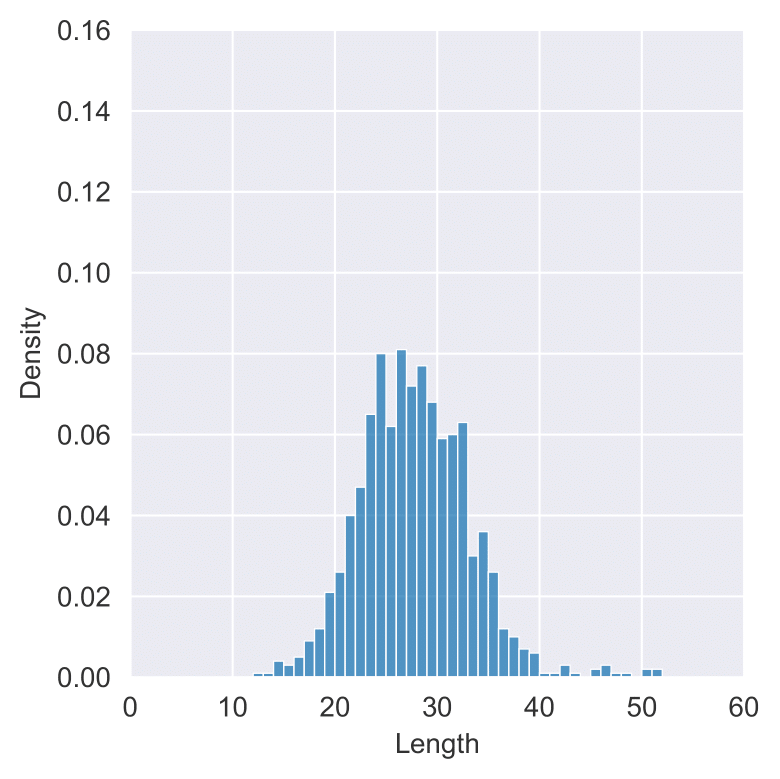}
         \caption{\protect\sa{} 
         }
         \label{fig:l-sa-en}
    \end{subfigure}
     \begin{subfigure}[b]{0.32\textwidth}
         \centering
         \includegraphics[width=\textwidth]{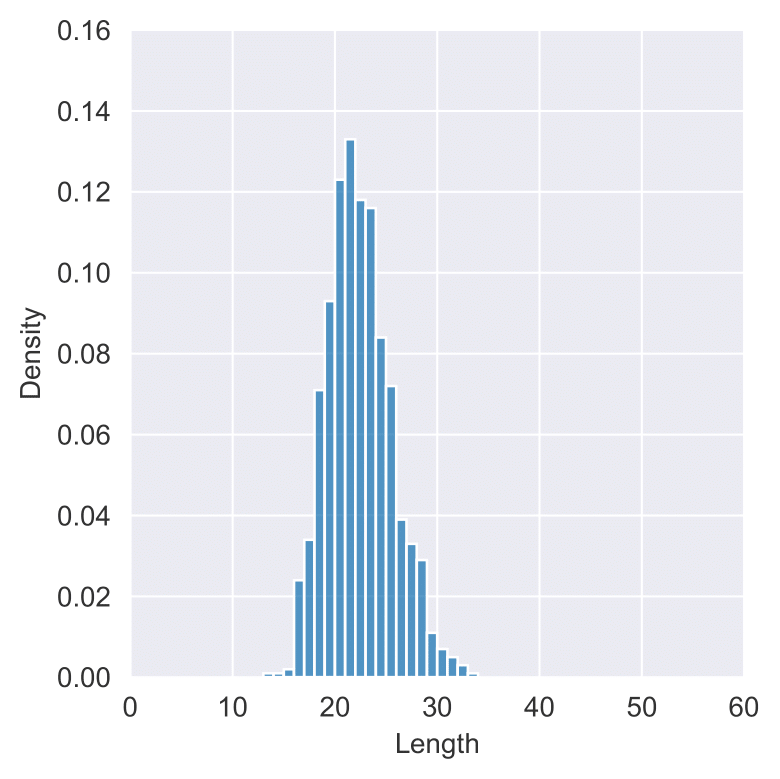}
         \caption{\protect\gptneol{} 
         }
         \label{fig:l-GPTneo_xl-en}
    \end{subfigure}
    \vspace{-.1cm}
    \caption{Length distribution of human poems (left), \protect\sa{} (middle) and \protect\gptneol{} (right) for English.}
    \label{fig:length}
    \vspace{-.3cm}
\end{figure*}

\paragraph{Length}
Table \ref{tab:length} displays the length related statistics. Figure \ref{fig:length} illustrates the length distribution of human written poems, \sa{} and \gptneol{} for English.

\begin{figure*}[!t]
    \centering
    \begin{subfigure}[b]{0.24\textwidth}
         \centering
         \includegraphics[width=\textwidth]{images/revision_june/rhymes/de_1DS_-___histogram.pdf}
         \caption{\protect\ds{} (de)}
    \end{subfigure}
    \hfill
    \begin{subfigure}[b]{0.24\textwidth}
         \centering
         \includegraphics[width=\textwidth]{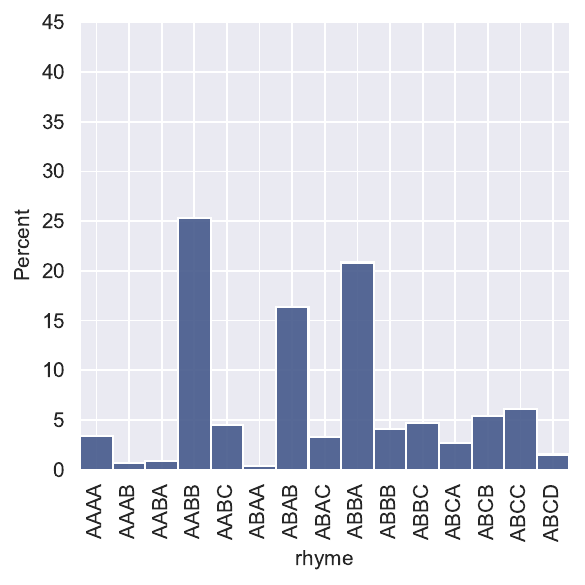}
         \caption{\protect\ds{} (en)}
    \end{subfigure}
    \hfill
     \begin{subfigure}[b]{0.24\textwidth}
         \centering
         \includegraphics[width=\textwidth]{images/revision_june/rhymes/de_2SA_histogram.pdf}
         \caption{\protect\sa{} (de)}
    \end{subfigure}
    \begin{subfigure}[b]{0.24\textwidth}
         \centering
         \includegraphics[width=\textwidth]{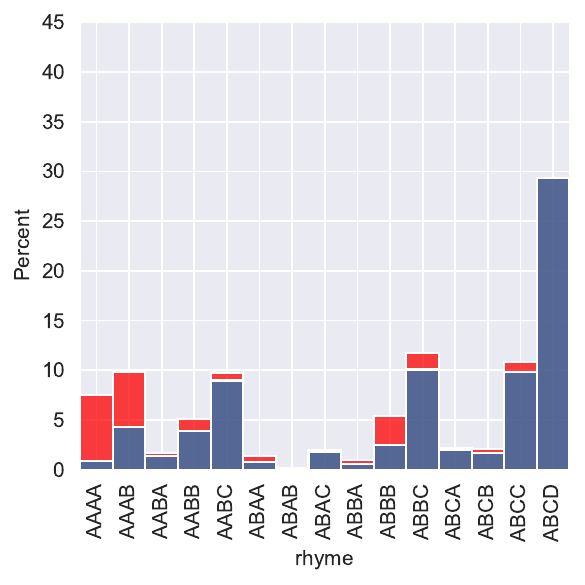}
         \caption{\protect\sa{} (en)}
    \end{subfigure}
    \caption{\yc{Distribution of rhyme schemes in the samples from \protect\ds{} and \protect\sa{} models for German and English.} 
    } 
    \label{fig:r-ds-sa}
\end{figure*}

\begin{figure*}[!h]
    \centering
    \begin{subfigure}[b]{0.325\textwidth}
         \centering
         \includegraphics[width=\textwidth]{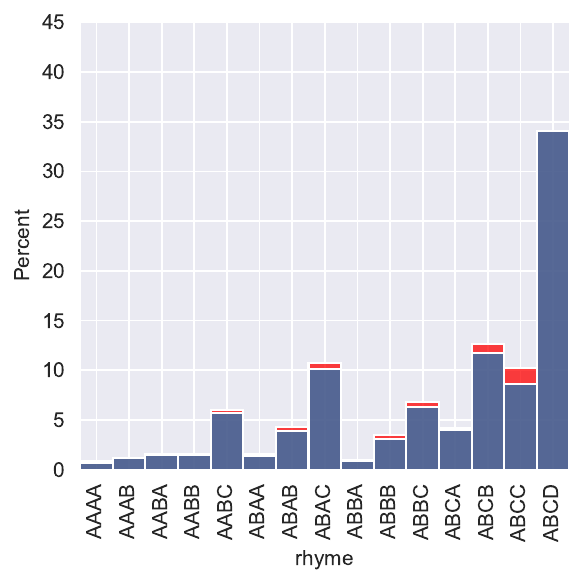}
         \caption{\protect\bygpts{}}
    \end{subfigure}
    \hfill
    \begin{subfigure}[b]{0.325\textwidth}
         \centering
         \includegraphics[width=\textwidth]{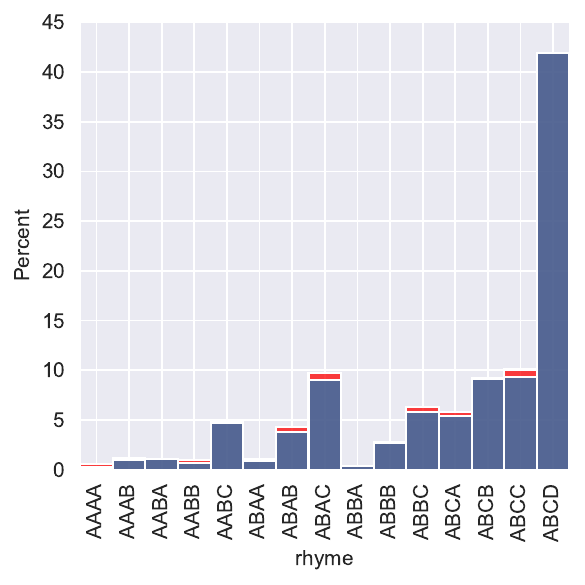}
         \caption{\protect\bygptl{}}
    \end{subfigure}
    \hfill
     \begin{subfigure}[b]{0.325\textwidth}
         \centering
         \includegraphics[width=\textwidth]{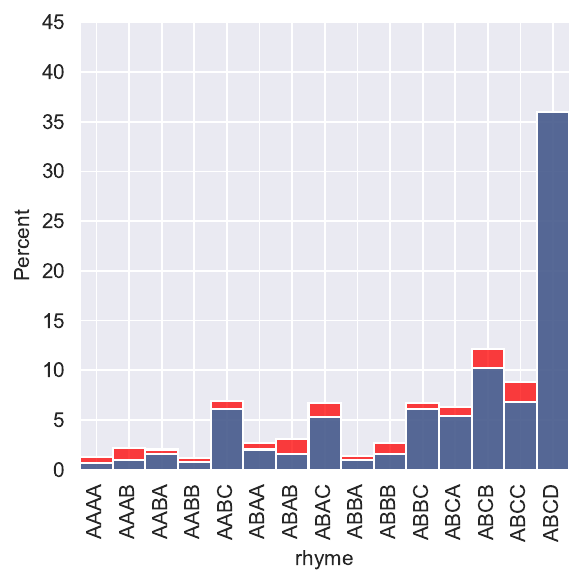}
         \caption{\protect\gpttwos{}}
    \end{subfigure}
    \begin{subfigure}[b]{0.325\textwidth}
         \centering
         \includegraphics[width=\textwidth]{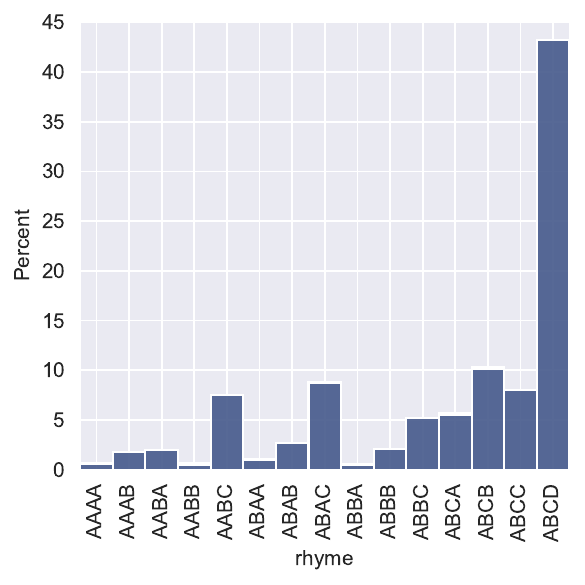}
         \caption{\protect\gpttwol{}}
    \end{subfigure}
    \begin{subfigure}[b]{0.325\textwidth}
         \centering
         \includegraphics[width=\textwidth]{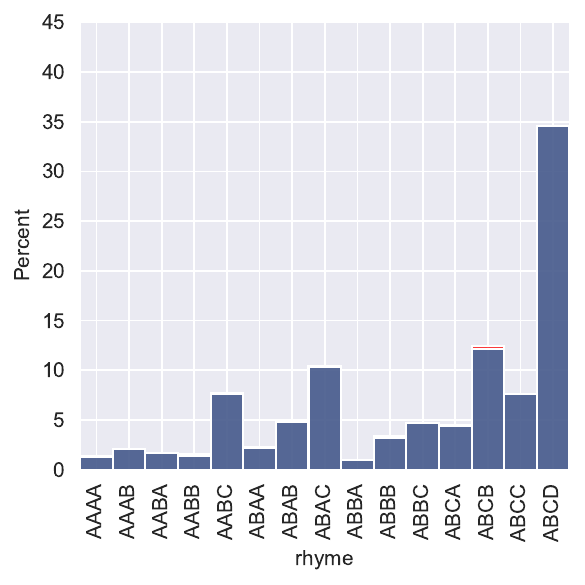}
         \caption{\protect\llamatwos{}}
    \end{subfigure}
    \hfill
    \begin{subfigure}[b]{0.325\textwidth}
         \centering
         \includegraphics[width=\textwidth]{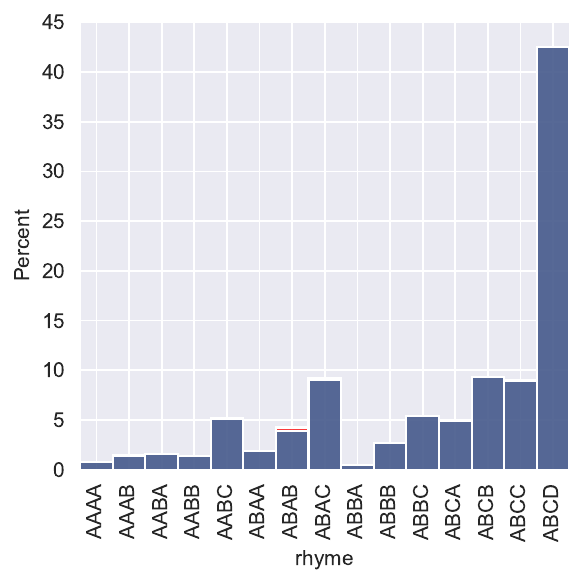}
         \caption{\protect\llamatwol{}}
    \end{subfigure}
    \hfill
     \begin{subfigure}[b]{0.325\textwidth}
         \centering
         \includegraphics[width=\textwidth]{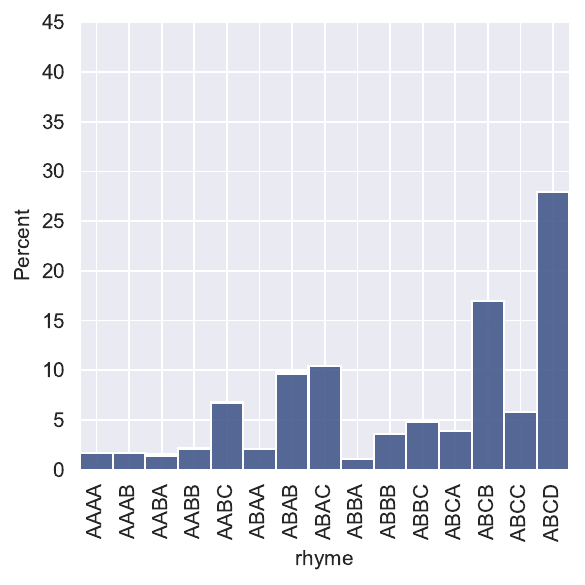}
         \caption{\protect\llamathr{}}
    \end{subfigure}
    \caption{Rhyme distribution plots for samples generated by \textbf{German unconditioned} large language models.} 
    \label{fig:r-nonpoetry-de}
\end{figure*}

\begin{figure*}[!h]
    \centering
    \begin{subfigure}[b]{0.325\textwidth}
         \centering
         \includegraphics[width=\textwidth]{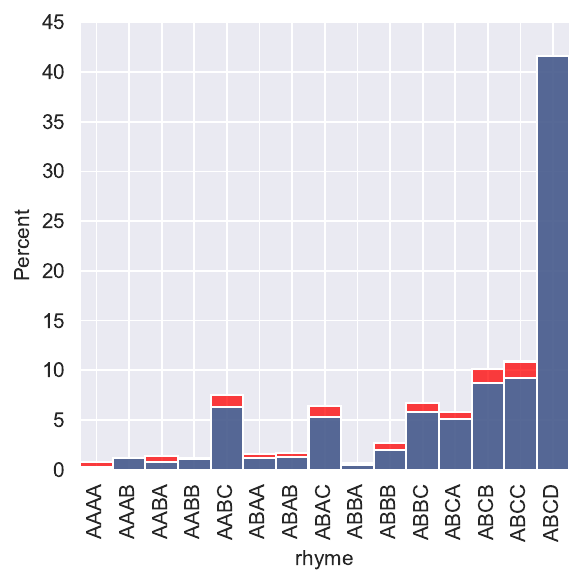}
         \caption{\protect\bygpts{}}
    \end{subfigure}
    \hfill
    \begin{subfigure}[b]{0.325\textwidth}
         \centering
         \includegraphics[width=\textwidth]{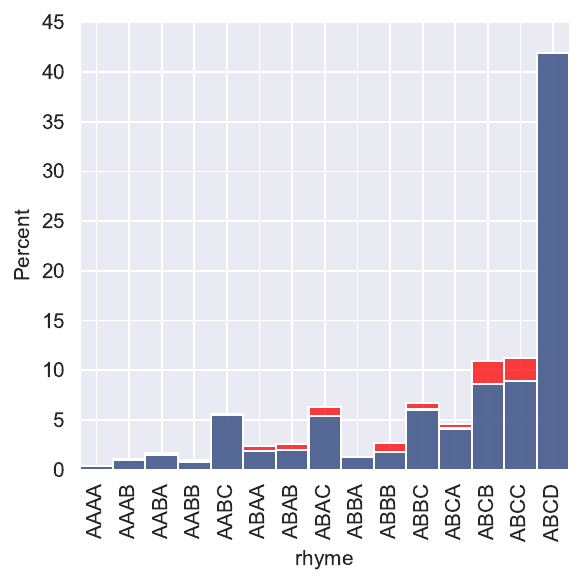}
         \caption{\protect\bygptl{}}
    \end{subfigure}
    \hfill
     \begin{subfigure}[b]{0.325\textwidth}
         \centering
         \includegraphics[width=\textwidth]{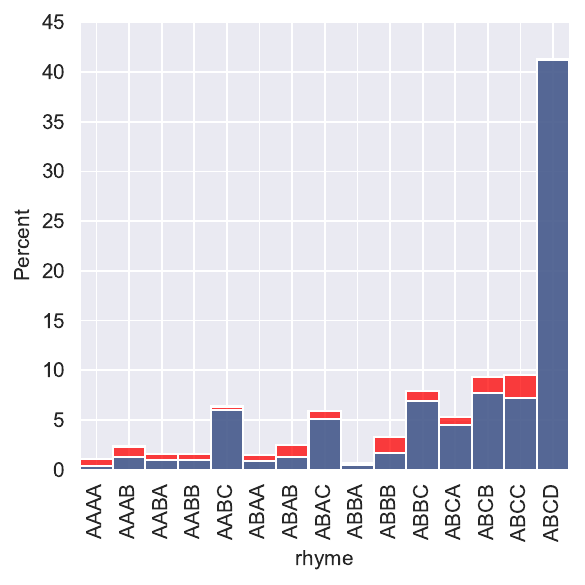}
         \caption{\protect\gpttwos{}}
    \end{subfigure}
    \begin{subfigure}[b]{0.325\textwidth}
         \centering
         \includegraphics[width=\textwidth]{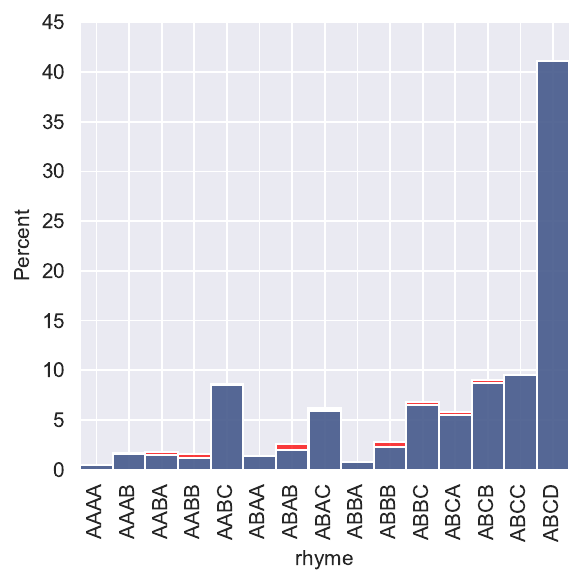}
         \caption{\protect\gpttwol{}}
    \end{subfigure}
     \begin{subfigure}[b]{0.335\textwidth}
         \centering
         \includegraphics[width=\textwidth]{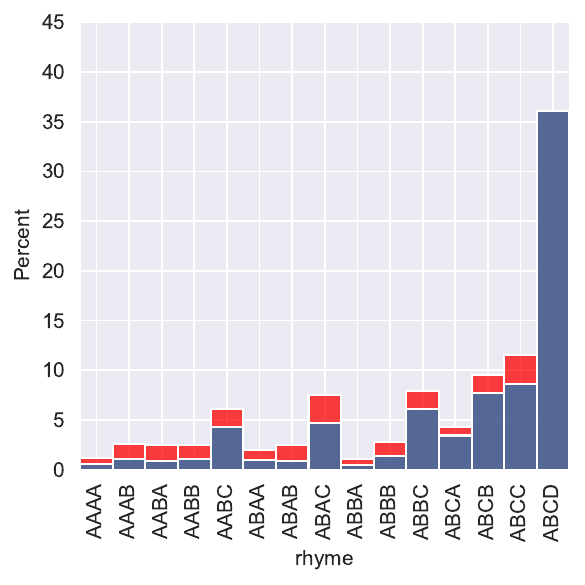}
         \caption{\protect\gptneos{}}
    \end{subfigure}
    \begin{subfigure}[b]{0.325\textwidth}
         \centering
         \includegraphics[width=\textwidth]{images/revision_june/rhymes/unconditioned/en_GPTNEO_L_unc_histogram.pdf}
         \caption{\protect\gptneol{}}
    \end{subfigure}
    \begin{subfigure}[b]{0.325\textwidth}
         \centering
         \includegraphics[width=\textwidth]{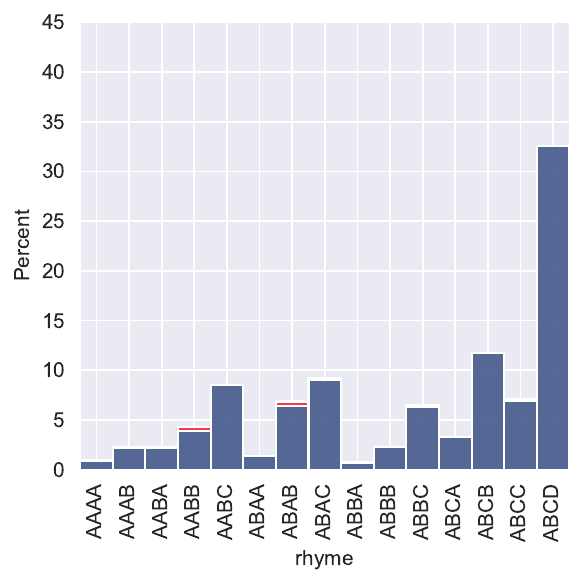}
         \caption{\protect\llamatwos{}}
    \end{subfigure}
    \hfill
    \begin{subfigure}[b]{0.325\textwidth}
         \centering
         \includegraphics[width=\textwidth]{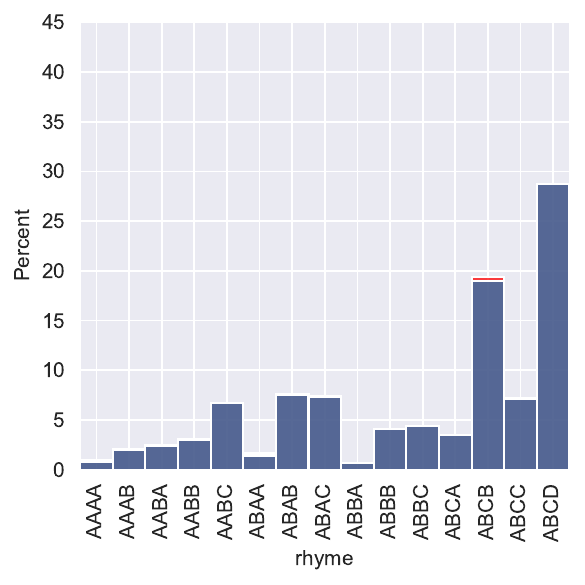}
         \caption{\protect\llamatwol{}}
    \end{subfigure}
    \hfill
     \begin{subfigure}[b]{0.325\textwidth}
         \centering
         \includegraphics[width=\textwidth]{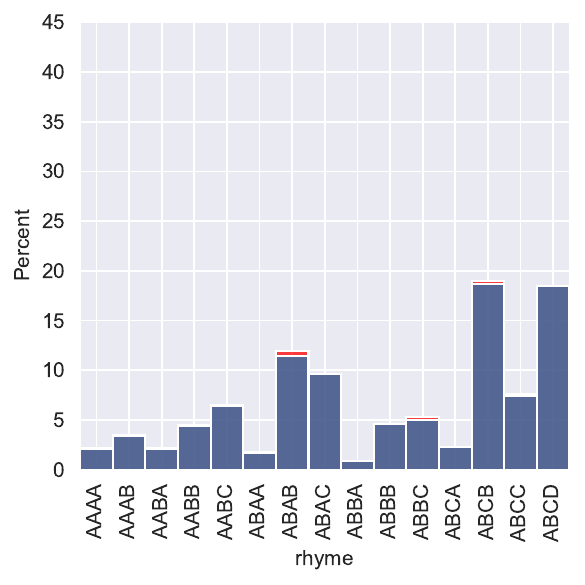}
         \caption{\protect\llamathr{}}
    \end{subfigure}
    \caption{Rhyme distribution plots for samples generated by \textbf{English unconditioned} large language models.} 
    \label{fig:r-nonpoetry-en}
\end{figure*}

\begin{figure*}[!h]
    \centering
    \begin{subfigure}[b]{0.325\textwidth}
         \centering
         \includegraphics[width=\textwidth]{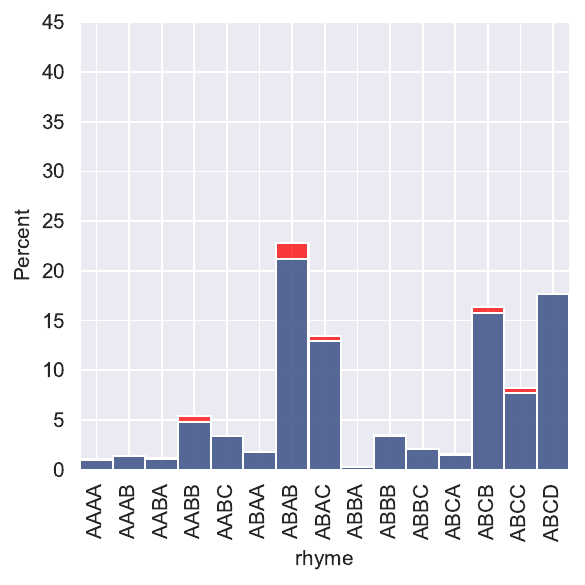}
         \caption{\protect\bygptsc{}}
    \end{subfigure}
    \hfill
    \begin{subfigure}[b]{0.325\textwidth}
         \centering
         \includegraphics[width=\textwidth]{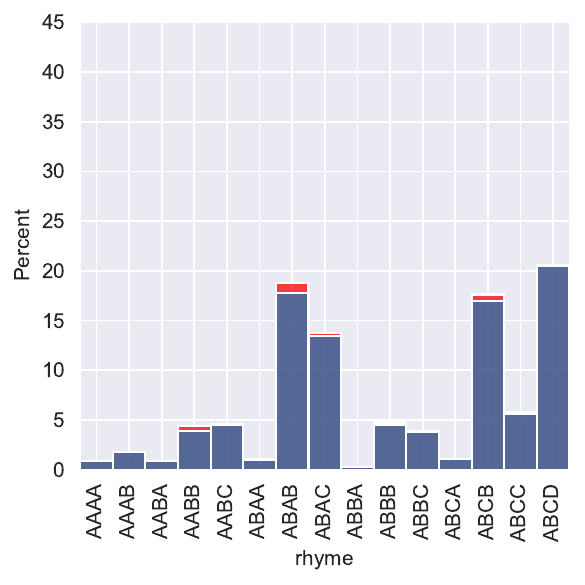}
         \caption{\protect\bygptlc{}}
    \end{subfigure}
    \hfill
     \begin{subfigure}[b]{0.325\textwidth}
         \centering
         \includegraphics[width=\textwidth]{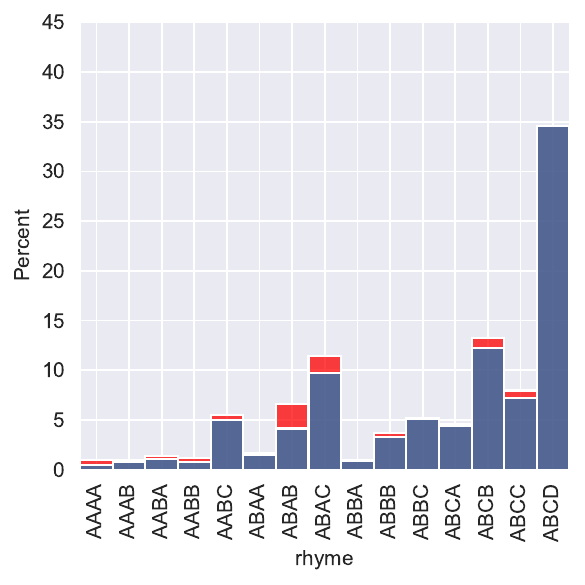}
         \caption{\protect\gpttwosc{}}
    \end{subfigure}
    \begin{subfigure}[b]{0.325\textwidth}
         \centering
         \includegraphics[width=\textwidth]{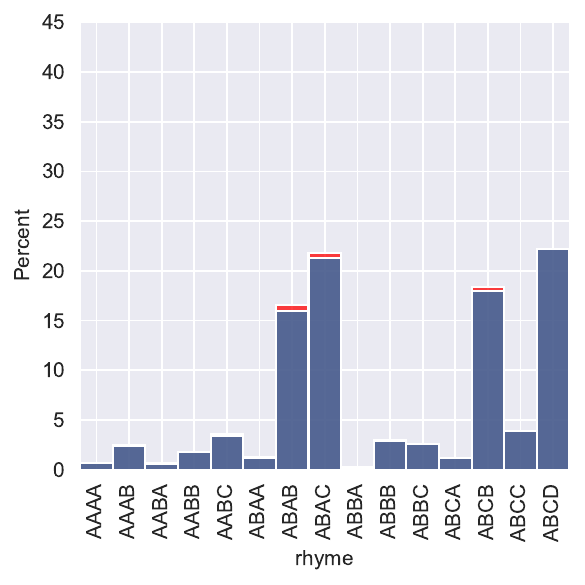}
         \caption{\protect\gpttwolc{}}
    \end{subfigure}
    \begin{subfigure}[b]{0.325\textwidth}
         \centering
         \includegraphics[width=\textwidth]{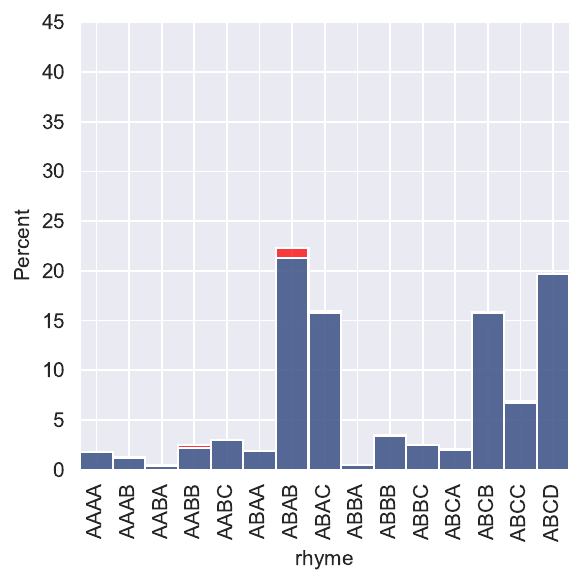}
         \caption{\protect\llamatwosc{}}
    \end{subfigure}
    \hfill
    \begin{subfigure}[b]{0.325\textwidth}
         \centering
         \includegraphics[width=\textwidth]{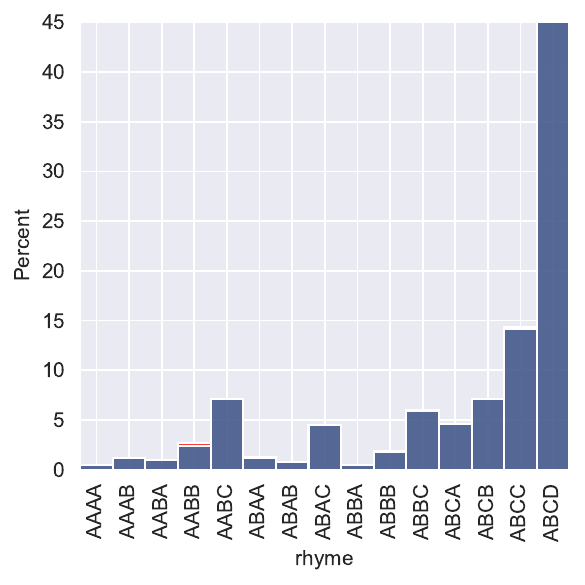}
         \caption{\protect\llamatwolc{}}
    \end{subfigure}
    \hfill
     \begin{subfigure}[b]{0.325\textwidth}
         \centering
         \includegraphics[width=\textwidth]{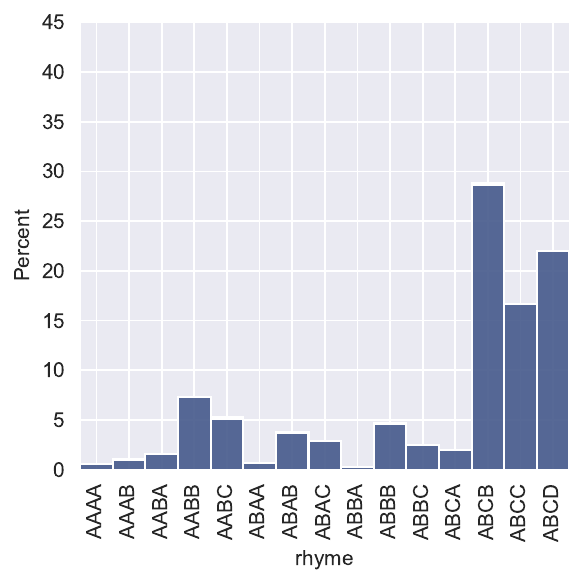}
         \caption{\protect\llamathrc{}}
    \end{subfigure}
    \caption{Rhyme distribution plots for samples generated by \textbf{German conditioned} large language models.} 
    \label{fig:r-poetry-de}
\end{figure*}

\begin{figure*}[!h]
    \centering
    \begin{subfigure}[b]{0.325\textwidth}
         \centering
         \includegraphics[width=\textwidth]{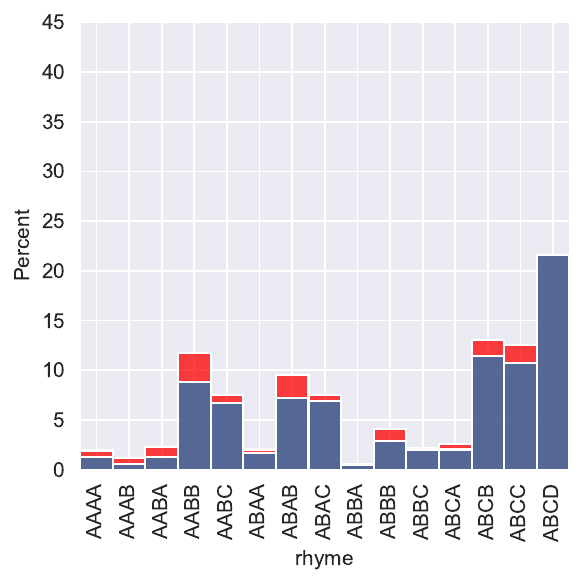}
         \caption{\protect\bygptsc{}}
    \end{subfigure}
    \hfill
    \begin{subfigure}[b]{0.325\textwidth}
         \centering
         \includegraphics[width=\textwidth]{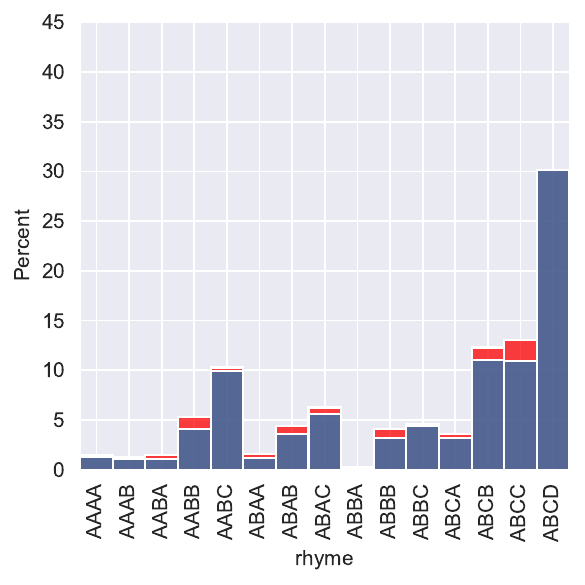}
         \caption{\protect\bygptlc{}}
    \end{subfigure}
    \hfill
     \begin{subfigure}[b]{0.325\textwidth}
         \centering
         \includegraphics[width=\textwidth]{images/revision_june/rhymes/conditioned/en_GPT2_S_con_histogram.pdf}
         \caption{\protect\gpttwosc{}}
    \end{subfigure}
    \begin{subfigure}[b]{0.325\textwidth}
         \centering
         \includegraphics[width=\textwidth]{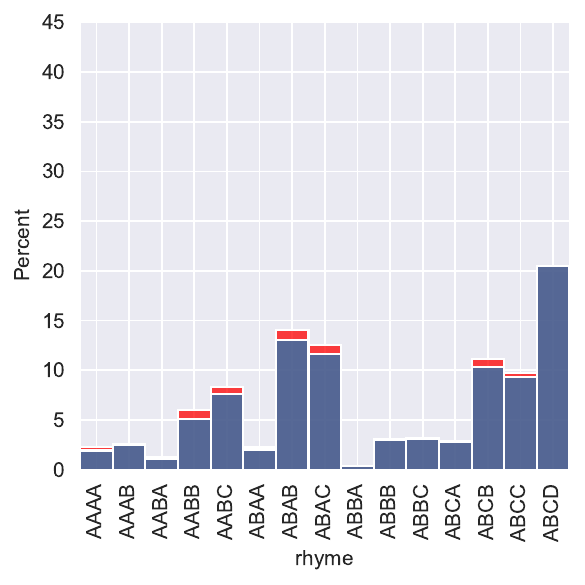}
         \caption{\protect\gpttwolc{}}
    \end{subfigure}
     \begin{subfigure}[b]{0.335\textwidth}
         \centering
         \includegraphics[width=\textwidth]{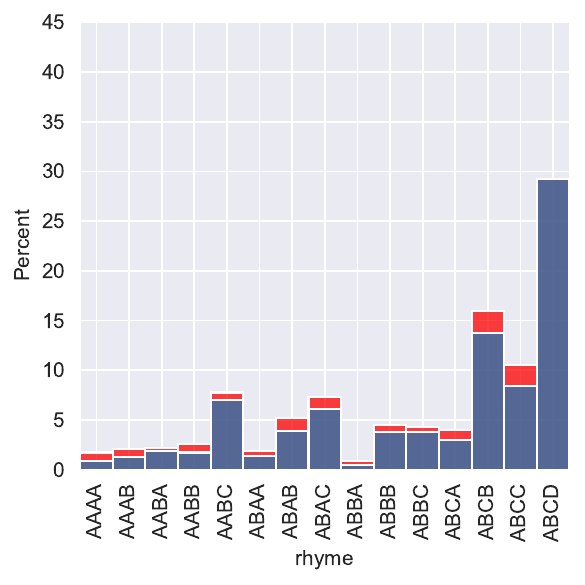}
         \caption{\protect\gptneosc{}}
    \end{subfigure}
    \begin{subfigure}[b]{0.325\textwidth}
         \centering
         \includegraphics[width=\textwidth]{images/revision_june/rhymes/conditioned/en_GPTNEO_L_con_histogram.pdf}
         \caption{\protect\gptneolc{}}
    \end{subfigure}
    \begin{subfigure}[b]{0.325\textwidth}
         \centering
         \includegraphics[width=\textwidth]{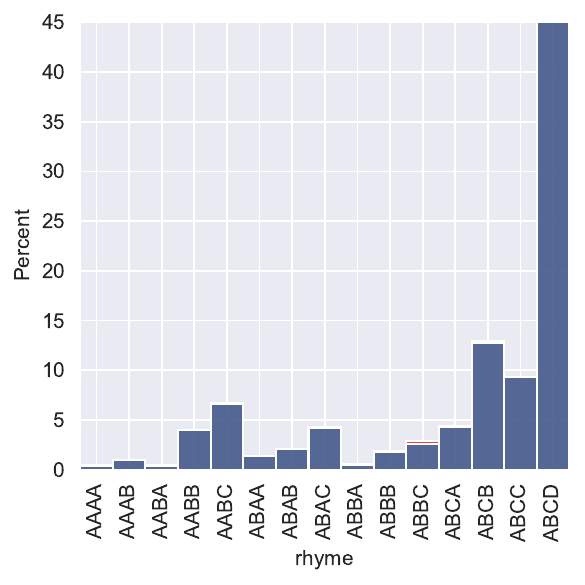}
         \caption{\protect\llamatwosc{}}
    \end{subfigure}
    \hfill
    \begin{subfigure}[b]{0.325\textwidth}
         \centering
         \includegraphics[width=\textwidth]{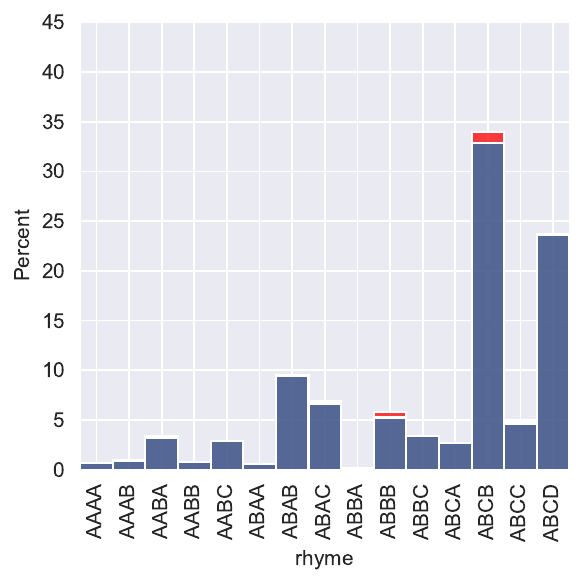}
         \caption{\protect\llamatwolc{}}
    \end{subfigure}
    \hfill
     \begin{subfigure}[b]{0.325\textwidth}
         \centering
         \includegraphics[width=\textwidth]{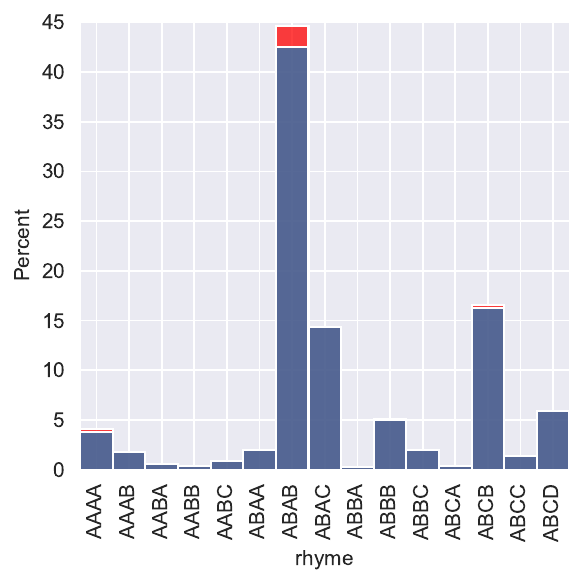}
         \caption{\protect\llamathrc{}}
    \end{subfigure}
    \caption{Rhyme distribution plots for samples generated by \textbf{English conditioned} large language models.} 
    \label{fig:r-poetry-en}
\end{figure*}

\paragraph{Rhyme}
Table \ref{tab:entropy} shows the entropy of the rhyme distributions in each sample as well as the distances of the distributions to that in the human data, measured by KL divergence. Figure \ref{fig:rhyme_en} demonstrates the human rhyme distribution as well as the best, worst, and an average fit distributions in terms of KL divergence.
Figures \ref{fig:r-ds-sa}, \ref{fig:r-nonpoetry-de}/\ref{fig:r-nonpoetry-en}, and \ref{fig:r-poetry-de}/\ref{fig:r-poetry-en} demonstrate the rhyme distributions for the poetry specific models, unconditioned and conditioned LLMs, respectively.

\begin{table*}[!h]
\centering
\begin{tabular}{@{}lcccc@{}}
\toprule
 & \multicolumn{2}{c}{\textbf{DE}} & \multicolumn{2}{c}{\textbf{EN}} \\ 
\cmidrule(lr){2-3}
\cmidrule(lr){4-5} 
Model      & Entropy & KL Divergence  & Entropy & KL Divergence  \\ \cmidrule(lr){1-3}
\cmidrule(lr){4-5} 
\textit{HUMAN}                      & 2.90    & 0.00                                & 3.10	&0.00	   \\ \cmidrule(lr){1-3}
\cmidrule(lr){4-5} 
\ds{}                      & 2.97    & \textbf{0.55}                                &  3.16&	0.48	     \\
\sa{}                      & 3.14    & \underline{1.43}                                &   3.22	&1.17	    \\ \cmidrule(lr){1-3}
\cmidrule(lr){4-5} 
\bygptl{}               & 2.89    & 1.23                                &  2.92&	1.08	     \\
\bygpts{}                  & 3.13    & 1.09                                &  2.91&	1.13     \\
\gpttwol{}               & 2.86    & 1.26                                  &  2.97	&1.06	   \\
\gpttwos{}                  & 3.16    & 1.13                                 &  2.99&	1.03	    \\
\gptneol{} &-&-& 2.80	&\underline{1.18}	 \\
\gptneos{} &-&-& 3.16&	0.96	 \\
\llamatwol{}              & 2.93    & 1.18                                &   3.24&	0.71	    \\
\llamatwos{}            & 3.18    & 1.04                                    &3.24	&0.71	   \\
\llamathr{}                  & 3.27    & 0.83                                &  3.45&	0.56     \\ \cmidrule(lr){1-3}
\cmidrule(lr){4-5} 
\bygptlc{}              & 3.17    & 0.67                                  &  3.22&	0.83	   \\
\bygptsc{}                   & 3.16    & 0.58                                &  3.38&	0.54     \\
\gpttwolc{}                & 2.98    & 0.99                                &    3.41&	0.61	   \\
\gpttwosc{}                   & 3.11    & 1.04                                  & 3.22	&0.85	   \\
\gptneolc{} &-&-& 3.43	&\textbf{0.45} \\
\gptneosc{} &-&-& 3.29	&0.83	 \\
\llamatwolc{}                  & 2.69    & 1.33                               &  2.89&	0.95	      \\
\llamatwosc{}              & 3.11    & 0.71                                &  2.67&	1.07     \\
\llamathrc{}                 & 2.98    & 1.06                                &  2.58	&0.94   \\ \bottomrule
\end{tabular}
\caption{Entropy and KL divergence of rhyme distributions. We bold the lowest and underline the highest KL divergence from human to model distributions.}
\label{tab:entropy}
\end{table*}

\paragraph{Best model}\label{app:best}

We rank the models for each dimension and then average the ranks across the five dimensions to determine the overall rankings. For dimensions with multiple metrics, such as the three memorization metrics (due to different evaluation levels) and the three lexical metrics (measuring local or global lexical diversity), we first rank the models according to each metric and then average these ranks to represent that dimension. For dimensions primarily based on distributions, we use metrics that measure the distance/similarity of their distributions from human data: KL divergence for rhyme and histogram intersection for length.
The results are shown in Table \ref{tab:rank_de} and \ref{tab:rank_en} for German and English respectively.

\begin{table*}[t]
\small 
\centering
\begin{tabular}{@{}llllcccccc@{}} 
\toprule
Language & Model    & Size & Conditioned & semantic & lexical & length & rhyme & memorization & avg\_rank \\ \midrule
de       & BYGPT5 & L    & TRUE        & 2.0      & 4.0     & 5.0    & 3.0   & 1.7          & 3.1       \\
de       & BYGPT5 & S    & TRUE        & 3.5      & 6.0     & 4.0    & 2.0   & 1.3          & 3.4       \\
de       & SA      & -    & -       & 1.0      & 2.7     & 1.0    & 16.0  & 2.0          & 4.5       \\
de       & DS      & -    & -       & 5.0      & 10.3    & 12.0   & 1.0   & 1.0          & 5.9       \\
de       & BYGPT5 & S    & FALSE       & 6.0      & 11.0    & 2.0    & 10.0  & 2.7          & 6.3       \\
de       & BYGPT5 & L    & FALSE       & 4.0      & 8.3     & 6.0    & 13.0  & 3.0          & 6.9       \\
de       & LLAMA3   & -    & FALSE       & 9.5      & 6.3     & 9.0    & 5.0   & 6.0          & 7.2       \\
de       & LLAMA3   & -    & TRUE        & 6.5      & 7.3     & 8.0    & 9.0   & 5.7          & 7.3       \\
de       & LLAMA2   & S    & TRUE        & 13.5     & 13.0    & 3.0    & 4.0   & 4.0          & 7.5       \\
de       & GPT2     & L    & TRUE        & 12.5     & 4.7     & 7.0    & 6.0   & 8.3          & 7.7       \\
de       & LLAMA2   & L    & FALSE       & 9.5      & 2.7     & 16.0   & 12.0  & 5.3          & 9.1       \\
de       & LLAMA2   & S    & FALSE       & 8.0      & 10.0    & 15.0   & 8.0   & 5.0          & 9.2       \\
de       & GPT2     & L    & FALSE       & 14.0     & 5.7     & 10.0   & 14.0  & 8.7          & 10.5      \\
de       & GPT2     & S    & TRUE        & 15.0     & 15.0    & 11.0   & 7.0   & 6.3          & 10.9      \\
de       & LLAMA2   & L    & TRUE        & 12.5     & 13.0    & 13.0   & 15.0  & 8.0          & 12.3      \\
de       & GPT2     & S    & FALSE       & 13.5     & 16.0    & 14.0   & 11.0  & 7.7          & 12.4   \\ \bottomrule  
\end{tabular}
\caption{Ranking of \textbf{German} models for each dimension, as well as the average ranks across all dimensions.}\label{tab:rank_de}
\end{table*}

\begin{table*}[t]
\centering
\small 
\begin{tabular}{@{}llllcccccc@{}}
\toprule
Language & Model    & Size & Conditioned & semantic & lexical & length & rhyme & memorization & avg\_rank \\ \midrule
en       & BYGPT5 & S    & TRUE        & 3.5      & 11.7    & 4.0    & 3.0   & 2.0          & 4.8       \\
en       & SA      & -    & -       & 1.0      & 4.0     & 1.0    & 19.0  & 1.0          & 5.2       \\
en       & BYGPT5 & L    & TRUE        & 2.0      & 9.7     & 5.0    & 9.0   & 1.7          & 5.5       \\
en       & DS      & -    & -       & 3.5      & 9.0     & 17.0   & 2.0   & 2.3          & 6.8       \\
en       & LLAMA2   & S    & FALSE       & 17.5     & 5.7     & 2.0    & 6.0   & 4.7          & 7.2       \\
en       & LLAMA3   & -    & TRUE        & 12.0     & 1.7     & 9.0    & 11.0  & 3.3          & 7.4       \\
en       & GPT2     & L    & TRUE        & 9.0      & 9.0     & 6.0    & 5.0   & 9.3          & 7.7       \\
en       & LLAMA2   & L    & TRUE        & 12.0     & 5.0     & 7.0    & 12.0  & 4.0          & 8.0       \\
en       & LLAMA2   & S    & TRUE        & 7.0      & 3.3     & 13.0   & 16.0  & 1.3          & 8.1       \\
en       & LLAMA3   & -    & FALSE       & 13.0     & 3.0     & 16.0   & 4.0   & 9.0          & 9.0       \\
en       & LLAMA2   & L    & FALSE       & 9.0      & 6.3     & 15.0   & 7.0   & 10.3         & 9.5       \\
en       & GPT2     & S    & TRUE        & 17.5     & 14.0    & 3.0    & 10.0  & 3.7          & 9.6       \\
en       & BYGPT5 & L    & FALSE       & 5.5      & 15.7    & 10.0   & 17.0  & 3.0          & 10.2      \\
en       & BYGPT5 & S    & FALSE       & 5.5      & 17.3    & 8.0    & 18.0  & 2.7          & 10.3      \\
en       & GPTNEO   & L    & TRUE        & 13.5     & 13.0    & 19.0   & 1.0   & 10.0         & 11.3      \\
en       & GPTNEO   & S    & TRUE        & 16.0     & 17.0    & 11.0   & 8.0   & 5.7          & 11.5      \\
en       & GPT2     & L    & FALSE       & 10.5     & 11.0    & 12.0   & 15.0  & 11.3         & 12.0      \\
en       & GPT2     & S    & FALSE       & 17.0     & 19.0    & 14.0   & 14.0  & 11.7         & 15.1      \\
en       & GPTNEO   & S    & FALSE       & 17.5     & 20.0    & 18.0   & 13.0  & 12.0         & 16.1      \\
en       & GPTNEO   & L    & FALSE       & 17.5     & 14.7    & 20.0   & 20.0  & 11.3         & 16.7     \\ \bottomrule
\end{tabular}
\caption{Ranking of \textbf{English} models for each dimension, as well as the average ranks across all dimensions.}\label{tab:rank_en}
\end{table*}

\begin{table*}[!ht]
\small
\setlength{\tabcolsep}{2pt}
\resizebox{\textwidth}{!}{%
\begin{tabular}{@{}l|ccccc|ccccc@{}}
\toprule
      & \multicolumn{5}{c|}{\underline{DE}}                             & \multicolumn{5}{c}{\underline{EN}}                                           \\
 \multirow{2}{*}{System}   & \multirow{2}{*}{HUMAN} & overall        & semantic & lexical     & rhyme & \multirow{2}{*}{HUMAN} & overall        & semantic & lexical     & rhyme           \\
      &       & (\bygptlc{}) & (\sa)        & (\sa)       & (\ds)     &       & (\bygptsc{}) & (\sa)       & (\llamathrc) & (\gptneolc{}) \\ \midrule
Best  & 12    & 2              & 0        & -       & 1     & 3     & 1              & 0        & 6           & 5              \\
Worst & 0     & 1              & 8        & -       & 6     & 2     & 2              & 11       & 0           & 0              \\
BWS   & \textbf{0.8}   & \underline{0.07}           & -0.53    & -       & -0.33 & 0.07  & -0.07          & -0.73    & \underline{\textbf{0.4}}         & 0.33           \\ \bottomrule
\end{tabular}
}
\caption{Best-worst scaling results of quality evaluation for human-written quatrains and quatrains generated by the most semantically, lexically, and rhythmically diverse systems.}\label{tab:quality_eval}
\end{table*}

\begin{table*}[!ht]
\centering
\begin{tabular}{@{}ll@{}}
\toprule
Quatrain                                                                                                                                                                                                                                              & System        \\ \midrule
\begin{tabular}[c]{@{}l@{}}sie lächelt, sprach doch: »ich bin\\ durch meine hand gefangen!\\ wir wollen diese liebe nicht verlangen,\\ und kommen zu dir angelangen.\end{tabular}                                                                   & \bygptlc \\\midrule
\begin{tabular}[c]{@{}l@{}}was werd' ich morgen tun?\\ ich könnt' ja nicht zu hause bleiben,\\ die nacht wird frieren,\\ der tag wird bald verschwinden.\end{tabular}                                                                               & \bygptlc \\\midrule
\begin{tabular}[c]{@{}l@{}}und sagt: was hat der mensch gebracht\\ was thut dir für die nacht\\ doch ist es halb, nicht schön zu sein\\ mein gott, ist andre ein\end{tabular}                                                                       & \ds   \\\midrule
\begin{tabular}[c]{@{}l@{}}hier wars, hier lag ich, auf der stelle,\\ in diesem veilchenvollen gras;\\ an diesem baum, bey dieser quelle,\\ da träumte mir vom jungen licidas!\end{tabular}                                                         & HUMAN         \\\midrule
\begin{tabular}[c]{@{}l@{}}drauf hebt sich ein gespräch von dessen wundern an;\\ da lächelt der vezier, und spricht zum suliman:\\ ich habe, großer held, bereits vor vielen jahren\\ die schwerste wissenschaft des orients erfahren.\end{tabular} & HUMAN         \\ \bottomrule
\end{tabular}
\caption{5 selected German quatrains rated as best in our human evaluation.}\label{tab:quatrain_example_de}
\end{table*}

\begin{table*}[!ht]
\centering
\begin{tabular}{@{}ll@{}}
\toprule
Quatrain                                                                                                                                                                               & System        \\ \midrule
\begin{tabular}[c]{@{}l@{}}it is the same old tune,\\ with its sweet, sad refrain;\\ but i'm not so sure of the new love's true name —\\ i have seen it before.\end{tabular}           & \llamathrc \\ \midrule
\begin{tabular}[c]{@{}l@{}}in this world, where we are born,\\ we see the same old face;\\ a little child at least has grown\\ to be our mother's grace.\end{tabular}                  & \gptneolc   \\ \midrule
\begin{tabular}[c]{@{}l@{}}thy brow is like the summer sky,\\ and all thy glances tell of spring;\\ the love that in thine eyes i see —\\ oh, sweetest song it ever sang!\end{tabular} & \llamathrc \\ \midrule
\begin{tabular}[c]{@{}l@{}}only when the night grows denser\\ march the bent monks one by one ,\\ singing to the sway of censer ,\\ kyrie — kyrie eleison !\end{tabular}               & HUMAN         \\ \midrule
\begin{tabular}[c]{@{}l@{}}a red rose burns upon his breast\\ where erst a white rose lay ;\\ above his fervent heart-throb pressed —\\ the red rose of to-day .\end{tabular}          & HUMAN         \\ \bottomrule
\end{tabular}
\caption{5 selected English quatrains rated as best in our human evaluation.}\label{tab:quatrain_example_en}
\end{table*}

\end{document}